\documentclass[10pt,twocolumn,letterpaper]{article}

\usepackage{iccv}
\usepackage{times}
\usepackage{epsfig}
\usepackage{titlecaps}
\usepackage{caption} 
\usepackage{svg}
\usepackage{pgf-pie}
\usepackage{array,multirow,graphicx}
\usepackage{verbatim}

\usepackage[pagebackref=true,breaklinks=true,letterpaper=true,colorlinks,bookmarks=false]{hyperref}

 \iccvfinalcopy 
\def\iccvPaperID{10227} 

\ificcvfinal\fi
\usepackage{amsmath}
\usepackage{bbm}
\usepackage{enumitem}
\usepackage{dirtree}
\usepackage{nicefrac}
\usepackage{tikz}
\usetikzlibrary{automata,calc,fit,positioning}
\usepackage{ulem}
\usepackage{forest}
\usepackage{wrapfig}
\usepackage{xcolor}
\usepackage{svg}
\usepackage{xparse}
\usepackage{xspace}
\usepackage[accsupp]{axessibility} 
\usepackage{xtab}
\usepackage{graphicx}
\usepackage{amsmath}
\usepackage{amssymb}
\usepackage{adjustbox} 
\usepackage{longtable}
\usepackage{pgf, tikz, pgfplots}
\usepackage{subcaption}
\usepackage{makecell}

\usepackage{threeparttable}

\usepackage{arydshln} 

\newcommand{\datasetabbr}{FACET\ }
\newcommand{\datasetabbrnospace}{FACET}

\newcommand{\datasourcenocite}{SA-1B}

\newcommand{\datasourcelong}{Segment Anything 1 Billion (SA-1B) \cite{segment_anything}}
\newcommand{\datasourcelongnocite}{Segment Anything 1 Billion (SA-1B)}

\newcommand{\gender}{perceived gender presentation\ }
\newcommand{\skintone}{perceived skin tone\ }
\newcommand{\age}{perceived age group\ }

\newcommand{\gendernospace}{perceived gender presentation}
\newcommand{\skintonenospace}{perceived skin tone}
\newcommand{\agenospace}{perceived age group}

\newcommand{\skintoneallcap}{Perceived Skin Tone}

\newcommand{\skintonecap}{Perceived skin tone }

\newcommand{\notab}{\hspace{-1.1em}}

\definecolor{Red}{rgb}{1,0,0}
\definecolor{Green}{rgb}{0,0.69,0}
\definecolor{Blue}{rgb}{0,0,1}
\definecolor{LightBlue}{rgb}{0,0.5,1}
\definecolor{veryLightBlue}{rgb}{0.85,0.98,1}
\definecolor{veryLightGreen}{rgb}{0.6,1,0.6}
\definecolor{Skin}{rgb}{1,0.71,0.69}
\definecolor{Grey}{rgb}{0.5,0.5,0.5}
\definecolor{LightGrey}{rgb}{0.6,0.6,0.6}
\definecolor{VeryLightGrey}{RGB}{219, 219, 219}
\definecolor{Black}{rgb}{0,0,0}
\definecolor{White}{rgb}{1,1,1}
\definecolor{brickred}{rgb}{0.8, 0.25, 0.33}
\definecolor{burntOrange}{RGB}{255,122,20}
\definecolor{navy}{RGB}{80, 74, 255}
\definecolor{teal}{RGB}{0, 123, 159}
\definecolor{aquamarine}{RGB}{51, 153, 255}
\definecolor{saffron}{RGB}{227, 170, 0}
\definecolor{purplePink}{RGB}{160, 89, 107}
\definecolor{xanadu}{RGB}{126, 145, 129}

\newcommand{\lightBlue}{\color{LightBlue}}

\newcommand{\orange}{\color{orange}}

\newcommand{\datasetsize}{31,702\ }
\newcommand{\peoplesize}{49,551\ }
\newcommand{\datasetsizemini}{32k\ }
\newcommand{\peoplesizemini}{50k\ }

\newcommand{\masksizemini}{69k\ }

\newcommand{\imagenetmini}{IN21k\ }

\newcommand{\attributenum}{13\ }

\newcommand{\numpersonclasses}{52\ }

\newcommand{\facetlink}{\url{https://facet.metademolab.com}}

\definecolor{m1}{HTML}{f6ede4}
\definecolor{m2}{HTML}{f3e7db}
\definecolor{m3}{HTML}{f7ead0}
\definecolor{m4}{HTML}{eadaba}
\definecolor{m5}{HTML}{d7bd96}
\definecolor{m6}{HTML}{a07e56}
\definecolor{m7}{HTML}{825c43}
\definecolor{m8}{HTML}{604134}
\definecolor{m9}{HTML}{3a312a}
\definecolor{m10}{HTML}{292420}

\newcommand{\mstone}{\color{m1} $\blacksquare$}
\newcommand{\msttwo}{\color{m2} $\blacksquare$}
\newcommand{\mstthree}{\color{m3} $\blacksquare$}
\newcommand{\mstfour}{\color{m4} $\blacksquare$}
\newcommand{\mstfive}{\color{m5} $\blacksquare$}
\newcommand{\mstsix}{\color{m6} $\blacksquare$}
\newcommand{\mstseven}{\color{m7} $\blacksquare$}
\newcommand{\msteight}{\color{m8} $\blacksquare$}
\newcommand{\mstnine}{\color{m9} $\blacksquare$}
\newcommand{\mstten}{\color{m10} $\blacksquare$}

\begin{document}

\title{FACET: Fairness in Computer Vision Evaluation Benchmark}

\author{
    Laura Gustafson\hspace{1.5em}
        Chloe Rolland\hspace{1.5em}
        Nikhila Ravi\hspace{1.5em}
        Quentin Duval\hspace{1.5em}
        Aaron Adcock\hspace{1.5em} \\
        Cheng-Yang Fu\hspace{1.5em}
        Melissa Hall\hspace{1.5em}
        Candace Ross\\
    Meta AI Research, FAIR\\
    {\tt facet@meta.com}
                                        }

\maketitle
\ificcvfinal\thispagestyle{empty}\fi

\begin{abstract}
    Computer vision models have known performance disparities across attributes such as gender and skin tone.
    This means during tasks such as classification and detection, model performance differs for certain classes based on the demographics of the people in the image.
    These disparities have been shown to exist, but until now there has not been a unified approach to measure these differences for common use-cases of computer vision models.
    We present a new benchmark named FACET ({\bf FA}irness in {\bf C}omputer Vision {\bf E}valua{\bf T}ion), a large, publicly available evaluation set of \datasetsizemini images for some of the most common vision tasks - image classification, object detection and segmentation.
    For every image in FACET, we hired expert reviewers to manually annotate person-related attributes such as \skintone and hair type, manually draw bounding boxes and label fine-grained person-related classes such as disk jockey or guitarist.
    In addition, we use \datasetabbr to benchmark state-of-the-art vision models and present a deeper understanding of potential performance disparities and challenges across sensitive demographic attributes.
   With the exhaustive annotations collected, we probe models using single demographics attributes as well as multiple attributes using an intersectional approach (\eg hair color and \skintonenospace).
    Our results show that classification, detection, segmentation, and visual grounding models exhibit performance disparities across demographic attributes and intersections of attributes. These harms suggest that not all people represented in datasets receive fair and equitable treatment in these vision
tasks. We hope current and future results using our benchmark will contribute to fairer, more robust vision models. FACET is available publicly at \facetlink.
   
\end{abstract}

\section{Introduction}\label{sec:intro}

\begin{figure}
\begin{center}
   \includegraphics[width=1.0\linewidth]{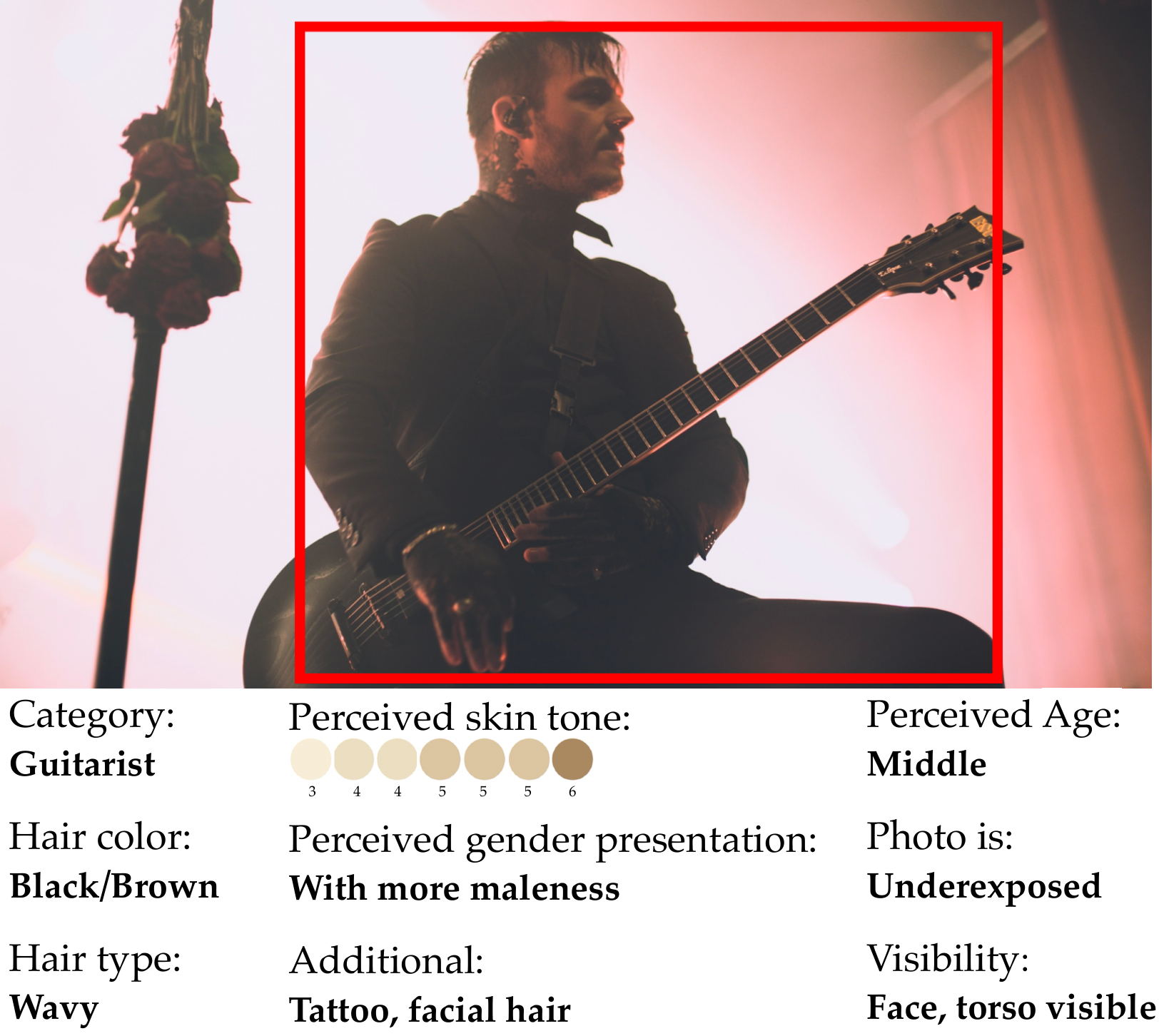}
\end{center}
\vspace{-1em}
   \caption{An example image and annotations from our dataset \datasetabbrnospace. Every image in \datasetabbr contains annotations from expert reviewers on the primary class, sensitive attributes including {\it\gendernospace, \skintonenospace}, and {\it\agenospace}, alongside additional visual attributes like \it{hair color and type}, \it{tattoos}, \it{\etc}}.
\label{fig:example}
\vspace{-1em}
\end{figure}

The ability of computer vision models to perform a wide range of tasks is due in no small part to large, widely used datasets.
These large-scale datasets containing millions of images often have image-level labels such as ImageNet \cite{imagenet_cvpr09} or object-level annotations found in datasets such as MS-COCO \cite{lin2014microsoft} or Open Images \cite{kuznetsova2020open}.
Annotations are also used at the person-level, in datasets such as CelebA \cite{liu2015celeba},
UTK-Faces \cite{zhifei2017cvpr}
and More Inclusive People Annotations (MIAP) \cite{MIAP}.
These person-level annotations in particular enable a more fine-grained analysis and evaluation of model performance across groups.
Prior work using these person-level annotations to evaluate model fairness has shown that vision models learn societal biases and stereotypes, which negatively impact performance and cause downstream harms \cite{steed2021image,zhao2017men,prabhu2020large,tong2020investigating,DBLP:journals/corr/abs-1711-11443}.
This makes fairness datasets particularly important as vision models continue to grow.

One weakness of existing fairness datasets is that they lack exhaustive and diverse demographic annotations that can support multiple vision tasks.
For instance, while Open Images More Inclusive People Annotations (MIAP) \cite{MIAP} can be used for classification and detection, the labels are not particularly diverse as only \gender and \age are labeled. Image-level class labels are also sparse, with an incomplete set of true positives and true negatives per image.
Another dataset, CelebA, contains many more person-level attributes but is primarily for face localization.
In addition, CelebA contains many subjective and potentially harmful attributes \eg attractive, big lips, chubby \cite{dooley2021comparing}.
These weaknesses can greatly impact our ability to perform more fine-grained fairness analyses.

In this paper, we present \datasetabbr (Fairness in Computer Vision Evaluation Benchmark), a large-scale evaluation benchmark with exhaustive annotations for \datasetsizemini images from 
\datasourcelong\ labeled across \attributenum person attributes and \numpersonclasses person classes.
The \attributenum attributes include examples such as \skintonenospace, hair type, \agenospace; the \numpersonclasses person classes include categories such as \textit{hairdresser} and \textit{reporter}. To ensure the annotations are both high quality and labeled by a diverse group of people, we used trained, expert annotators sourced from several geographic regions (North and Latin America, Middle East, Africa, East and Southeast Asia).

\begin{table}

    \begin{center}
    \scalebox{0.9}{
   \hspace*{-0.5em}\begin{tabular}{c|l}

         \textit{Size} & 
         -- \datasetsizemini images, \peoplesizemini people\\\hline
        \multirow{3}{*}{\parbox{5em}{\centering\textit{Evaluation\\Annotations}}}
        &
        -- 52-person related classes\\
        &
        -- bounding boxes around each person \\
        & 
        \parbox{18em}{-- person/hair/clothing labels for \masksizemini masks}\vspace*{0.5ex} \\\hline
         \multirow{3}{*}{\parbox{5em}{\centering\textit{Protected\\Groups}}}
         & -- \skintone \\
         & -- \age \\
         & -- \gender \\\hline

         \multirow{3}{*}{\parbox{5em}{\centering\textit{Additional\\Person Attributes}}}
         & -- {hair}: color, hair type, facial hair\\
         & -- {accessories}: headscarf, face mask, hat\\
         & -- {other}: tattoo\\\hline
         
         \multirow{2}{*}{\parbox{5.5em}{\centering\textit{Miscellaneous\\Attributes}}}
         &
         \multirow{2}{*}{lighting condition, level of occlusion} \\
         & \\
    \end{tabular}
    }
    \end{center}

\caption{Statistics on size of FACET and person annotations including labels for classification (\eg soldier, teacher) and attributes such as hair color and \skintonenospace.}
\label{fig:example}
\end{table}

FACET enables a deeper analysis of potential fairness concerns and model biases for specific demographic axes. We can explore questions such as: 1) \textit{Are models better at classifying people as skateboarder when their \gender has more stereotypically male attributes?} 2) \textit{Are open-vocabulary detection models better at detecting backpackers who are perceived to be younger?} 3) \textit{Do standard detection models struggle to detect people whose skin appears darker?} 4)
\textit{Are these problems magnified when the person has coily hair compared to straight hair?} 5) \textit{Do performance discrepancies differ across the detection and segmentation tasks?}
These questions illustrate a few examples of how model biases can be explored at a deep, intersectional level using the exhaustive annotations in \datasetabbrnospace.
We use \datasetabbr to evaluate multiple state-of-the-art vision models to understand their fairness on demographic attributes (\gendernospace, \skintonenospace, \agenospace) as well as their existing demographic biases. 
\datasetabbr is publicly available at \facetlink.
.

Our contributions include:
\begin{enumerate}[noitemsep,nolistsep]
    \item[--] our new publicly available fairness benchmark \datasetabbrnospace, containing \datasetsizemini images from \datasourcelong,  manually annotated with demographic and additional visual attributes labels by expert annotators
    \item[--] \numpersonclasses person-related class labels and manually drawn bounding boxes for every annotated person in every image (\peoplesizemini total people)
    \item[--] person, clothing or hair labels for \masksizemini masks
    \item[--] a benchmark for using \datasetabbr to compare different models, showing quantitative results and qualitative analyses on existing vision models using \datasetabbrnospace
\end{enumerate}

\textbf{\datasetabbr is an evaluation-only benchmark. Using any of the annotations for training is strictly prohibited.}

\section{Related Work}

\newcommand{\mtnote}[1]{\textsuperscript{\TPTtagStyle{#1}}}
\begin{table*}[h!tb]
\begin{threeparttable}
  \resizebox{\textwidth}{!}{
\begin{tabular}{c | r r r r r | r r r r r r | l}
    \it Dataset
    & \multicolumn{5}{c|}{\it Dataset Size}
    & \multicolumn{6}{c}{\it Apparent or Self-Reported Attributes }
    & \multicolumn{1}{c}{\it Task} \\
    & \parbox{3em}{\centering \#/people}
    & \parbox{3em}{\centering \#/images}
    & \parbox{3em}{\centering \#/videos}
    & \parbox{3em}{\centering \#/boxes}
    & \parbox{3em}{\centering \#/masks}
    & gender
    & age
    & skin tone
    & race
    & lighting
    & additional & \\ \hline
    
    UTK-Face\cite{zhifei2017cvpr}& 20k & 20k & -- & -- & -- & Yes & Yes & No & Yes & No & No & -- \\ \hline
    
    FairFace\cite{karkkainen2021fairface} &108k & 108k & -- & -- & -- & Yes & Yes & No & Yes & No & No & -- \\  \hline
    
    Gender Shades\cite{buolamwini2018gender} &1.2k & 1.2k & -- & -- & -- & Yes & Yes & Yes & No & No & No & -- \\ \hline
    
    OpenImages MIAP\cite{MIAP}&454k & 100k & -- & 454k & * & Yes & Yes & No & No & No & No & C\mtnote{*}DS\mtnote{*}\\ \hline   
    \cite{BDD100k_skintone} annotations for BDDK 100k \cite{bdd100k} & 16k & 2.2k & -- & 16k & * & No & No & Yes & No & Yes & No & DS\mtnote{*} \\ \hline
    
    \cite{zhao2021understanding} annotations for COCO \cite{lin2014microsoft} & 28k &  16k & -- & 28k & 28k & Yes & No & Yes & No & No & No & C\mtnote{*}DS \\ \hline
    
    Casual Conversations v1\cite{DBLP:journals/corr/abs-2104-02821}& 3k & N/A & 45k & -- & -- & Yes & Yes & Yes & No & Yes & Yes & -- \\ \hline

    Casual Conversations v2 \cite{ccv2} & 5.6k & N/A & 26k & -- & -- & Yes & Yes & Yes & No & Yes & Yes & -- \\ \hline   
    Ours – FACET & \peoplesizemini & \datasetsizemini & -- & \peoplesizemini & \masksizemini & Yes & Yes & Yes & No & Yes & Yes & CDS  \\  \hline
    
  \end{tabular}
  }
  \begin{tablenotes}\footnotesize
 \item[* ] represents tasks/annotations that are not included in the fairness portion of the dataset, but are included in the overall dataset.\\  e.g COCO has been used for multi-class classification \cite{zhao2017men,DirectionalBiasAmp}
\end{tablenotes}
\caption{Tasks and attribute annotations comparing existing datasets to FACET. These existing datasets were designed for fairness evaluations for other use cases, which is not to suggest that they are limited in use. The \textit{tasks} (CDS) considered are \textbf{C}lassification of an image, \textbf{D}etection of a person or person-related objects, \textbf{S}egmentation of a person or person-related objects. For classification, we do not include the classification task of classifying protected or non-protected attributes of a person. For attributes, \datasetabbr does not include race as it is not a visually salient category, exacerbates bias \cite{khan2021one} and misclassification has been shown to cause emotional distress \cite{doi:10.1177/000312240707200505}. Bounding boxes are denoted as boxes.}
\label{table:datasets_all} 
\end{threeparttable}
\end{table*}

Vision datasets that are annotated with apparent or self-reported demographic attributes are frequently used for studying model fairness.
Table \ref{table:datasets_all} compares FACET to other annotated datasets.

\hspace*{-1em}\textbf{Classification}
Datasets such as
\cite{liu2015celeba,zhifei2017cvpr,buolamwini2018gender,karkkainen2021fairface} are used to evaluate the gender and skin tone disparities in face recognition\footnote{We retain the same language used in the original papers, which is based on gender labels of the datasets that were audited.}. 
Gender Shades \cite{buolamwini2018gender}, for instance, showed that gender classification systems perform significantly worse on females compared to males and on darker skin compared to lighter skin using labels from annotators.
These datasets cannot be used for tasks outside of facial recognition, e.g. object detection or image classification.
Casual Conversations \cite{DBLP:journals/corr/abs-2104-02821} is a dataset used for videos; this dataset was used to highlight disparities across gender and skin tone for the Deep Fake Detection challenge \cite{DBLP:journals/corr/abs-2006-07397}.
Geographic and income-based disparities have been evaluated as well \cite{de2019does,sudhakar2023icon,gustafson2023pinpointing},
 most commonly with the
DollarStreet dataset \cite{dollarstreet,rojas2022the}.

\hspace{-1em}\textbf{Detection/Segmentation}
\cite{zhao2017men} generated gender annotations via captions for MS-COCO.
\cite{BDD100k_skintone} annotated a subsection of pedestrians in BDD100k \cite{bdd100k} for the task of pedestrian detection and found higher performance for lighter skin tones.
However, these demographic annotations are often noisy and either lack annotator training or lack annotators altogether and are instead approximated from captions.
More Inclusive Annotations for People (MIAP) \cite{MIAP}, which is a subset of Open Images \cite{kuznetsova2020open}, is dataset that does focus on high quality, more complete person-level demographic annotations for bounding boxes.
While MIAP is similar to \datasetabbrnospace, it only has annotations for \gender and \agenospace.
\datasetabbr has far more exhaustive annotations spanning far more attributes.
We have \attributenum attributes, including demographic attributes (\gendernospace, \agenospace, and \skintonenospace) as well as additional attributes such image quality (lighting and occlusion) and physical presentation (\eg hair type, accessories, tattoos, etc.). 

\hspace{-1.1em}\textbf{Best Practices}
Audits of popular computer vision datasets have found gender artifacts \cite{meister2022gender}, a lack of geographic diversity \cite{shankar2017no}, malignant stereotypes and NSFW content \cite{crawford_paglen_2019, prabhu2020large, birhane2021multimodal}.
To combat these issues, there has been significant research about dataset development including tools \cite{wang2022revise}, best practices for creating datasets \cite{koch2reduced, scheuerman2021datasets, andrews2023ethical, hutchinson2021towards, paullada2021data, ccv2, jo2020lessons, 49953, denton2020bringing, 10.1145/3392866} and designing annotation tasks with crowdworkers \cite{denton2021whose, daniel2018quality}. 
A large body of work also explored how researchers should document the intended use and considerations made when developing models \cite{mitchell2019model}, datasets \cite{gebru2021datasheets, pushkarna2022data, holland2018dataset, miceli2021documenting} and crowdsourced annotation tasks \cite{D_az_2022}.

\section{Benchmark Method}\label{bench-method}

The goal of our benchmark is to evaluate and analyze how vision models perform across different demographic and additional attributes for different categories of people.
This analysis requires (1) images that contain people with a diverse set of attributes and (2) images that contain people matching a variety of person-related categories. We focus on person-related categories such as occupations of people or person-related past-times, \eg \textit{doctor, basketball player, student, backpacker}, etc.
We prioritized a diverse set of categories for a more thorough analysis.

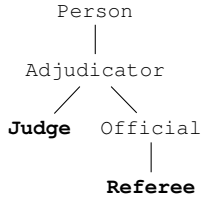
\begin{figure}[htb]
    \centering
    \footnotesize{
    \begin{forest}
    [
    \texttt{Person}, calign=child,calign child=3
        [\texttt{Adjudicator}
            [\texttt{\textbf{Judge}}]
            [\texttt{Official}
                [\texttt{\textbf{Referee}}]
            ]
        ]
    ]
    \end{forest}
    }
    \caption{
        Example of WordNet tree structure relating the FACET classes 
        \texttt{referee} and \texttt{judge} to the Person node.
    }
\label{fig:wn_ex}
\end{figure}

To generate the list of person-related categories, we use WordNet \cite{miller1995wordnet}, which is a hierarchical database of language concepts.
Each language concept is a single node in the hierarchy.
For instance, the concept \textit{apple} is a node with parent \textit{edible fruit}.
We take the language concept \texttt{person} and treat all of its children as potential categories, following \cite{TFD}. We filter out offensive synsets noted in \cite{TFD} and \cite{crawford_paglen_2019}, generating 1,239 candidate synsets.
We trim this list to \numpersonclasses categories using the considerations below. Figure \ref{fig:wn_ex} shows a sample WordNet tree structure for two classes in FACET, and 
Figure \ref{fig:wordnet_facet} in Appendix \ref{app:benchmark_decisions} shows the full hierarchy.

\paragraph{Connection to ImageNet-21k (IN21k)}
To ensure consistency with existing concepts used for computer vision model evaluation, we require our categories to overlap with the taxonomy of the widely used ImageNet-21k (IN21k) dataset \cite{ridnik2021imagenetk}.
This approach has been used to select object classes by other datasets \cite{barbu2019objectnet} and follows previous work \cite{goyal2022fairness, singh2022revisiting}.
This means models trained with IN21k can be evaluated out-of-the-box on \datasetabbrnospace.
\imagenetmini is a long-tailed dataset, meaning many classes have very few images.
We exclude categories with $<500 $ examples similar to \cite{ridnik2021imagenetk} to ensure that models pre-trained on IN21k will transfer seamlessly to our evaluation set.

\paragraph{Concept Selection} IN21k has overlapping classes with varying levels of specificity (\ie \texttt{surgeon} is a subcategory of \texttt{doctor}).
Following \cite{apple}, we include classes with roughly the same ``basic level." Using their findings of relative feature importance for classifying ``basic level", we limit the depth in the WordNet hierarchy from the \texttt{person} node to 5, as a proxy for level of specificity.
To alleviate ambiguity, we focus primarily on occupation/trade, sports, art, and leisure related categories of people.
This leaves us with \numpersonclasses categories.
Our final list of concepts is shown in Table~\ref{table:class_all} in the Appendix.

\section{Attribute Selection: Demographic and Additional Visual Attributes}

FACET includes both demographic attributes 
and additional visual attributes.
These exhaustively labeled, manually annotated attributes for all images in the dataset allow for evaluation of model performance and robustness at a fine-grained level.
For example, we can investigate potential biases associated with a single attribute as well as at the intersection of multiple attributes.
\textit{Intersectionality} is a framework exploring how multiple attributes can actually magnify societal biases \cite{crenshaw1989demarginalizing}; the exhaustive attributes in \datasetabbr means we can explore intersectional harms with respect to model fairness as well.
Examples questions we can explore include ``Do models struggle to classify people with tattoos?" (single attribute) and  ``Do models perform better for people with curly hair who appear to have perceived lighter skin tones than to those with perceived darker skin tones?" (intersection of attributes).
See Appendix \ref{app:data_stats} for the full list of attributes and their distributions.

\subsection{Demographic Attributes}\label{sec:demographic_attr}

\begin{figure}[htb]
\begin{center}
   \includegraphics[width=0.8\linewidth]{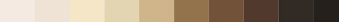}
\end{center}
\vspace*{-1em}
   \caption{Monk Skin Tone Scale\cite{Monk_2019}, an inclusive scale that includes 10 different skin tones.}
\label{fig:mst}
\end{figure}

\paragraph{\titlecap{\skintone}} 
The Monk Skin Tone Scale \cite{Monk_2019}, shown in Figure \ref{fig:mst}, was developed specifically for the computer vision use case.
We intentionally use the Monk Skin Tone scale over the Fitzpatrick skin type \cite{fitzpatrick1975soleil}, which was developed as means for determining one's likelihood of getting sunburn and lacks variance in darker skin tones \cite{DBLP:journals/corr/abs-2106-11240, 10.1111/bjd.19932}. Fitzpatrick skin type has been shown to be unreliable for image annotation \cite{10.1145/3555634}.

\textit{Skin tone as a spectrum.}
 Skin tone is difficult to annotate\footnote{Studies show even annotating one's own skin tone is difficult \cite{self-report-fitzpatrick}.} and can vary based on the lighting of the photo \cite{kakumanu2007survey}. For this reason, we annotate skin tone as a spectrum. We gather annotations from three different annotators, allowing annotators to select as many skin tone values they feel best represent the person. This gives us a distribution over various skin tones.
\textit{We note that \skintone is not a proxy for race or ethnicity, and should not be used as such.}

\paragraph{\titlecap{\gender}} Annotators select whether they perceive a person as having more stereotypically female attributes, having more stereotypically male attributes, or having attributes outside of the gender binary. We annotate \gender instead of gender, as gender cannot be determined purely from an image; attempting to do so can be harmful to groups who are misgendered \cite{10.1145/3173574.3173582}. A more thorough discussion is in Section \ref{sec:limitations}. 
\paragraph{\titlecap{\age}}
We have three distinct \age groups -- \textit{younger} are people perceived to be under 25 years old; \textit{middle} are people perceived to be between 25-65 years old and \textit{older} are people perceived to be over 65 years old. This follows the recommendations of \cite{ccv2} which matches the United Nation's breakdown of age \cite{UN}, but we collapse \texttt{Adults 25-40} and \texttt{Middle-age Adults 41-65} into one category \textit{middle}.
While it is impossible to tell a person's true age from an image, these numerical ranges are a rough guideline to delineate each \agenospace.

\subsection{Additional Attributes}\label{addn_attrs}

\paragraph{Hair color and hair type} Because conv-nets are shown to recognize patterns and textures \cite{gatys2015texture, gatys2017texture, geirhos2018imagenettrained, brendel2019approximating} and hair types represent a range of different textures, we annotate the hair color and hair type.\nolinebreak

\paragraph{Perceived lighting.}  Annotators labeled the lighting condition \textit{on the person}. This annotation is important in part because it heavily impacts perceived skin tone \cite{kakumanu2007survey}. 
These annotations can also guide the difficulty of the classification/detection problem, as models have been shown to have robustness vulnerabilities with respect to brightness \cite{hendrycks2018benchmarking, idrissi2022imagenet}.
\paragraph{Additional attributes.} We also annotate additional items relating to a person's appearance, using the recommendations of \cite{ccv2}. We condense the recommendations to the following list. 
These are \textbf{
facial hair,
head scarf \footnote{The motivation for this annotation is from a finding of \cite{singh2022revisiting} that the concept \textit{hijab} is predicted far more frequently for images with perceived lighter skin tones in UTK-Faces \cite{zhifei2017cvpr} than for those with perceived darker skin tones. It is unknown if this is a source of bias, as it is unknown whether or not there was a \textit{hijab} in the photo.},
hat/cap,
eyewear (eyeglasses/sunglasses),
face masks \footnote{Many images in FACET include more face masks than prior works, such as ImageNet, due to the COVID-19 pandemic.}, tattoos and a person's visibility
}.
\section{Annotation Method}

\subsection{Data Preprocessing}
\datasetabbr is composed of images from \datasourcelongnocite.
We preprocessed the dataset to create a rough pool of relevant examples (with a focus on high recall) before beginning the annotation process.
We use caption and tags for each image to create a candidate pool of images to annotate. First, for each of the \numpersonclasses selected categories, we created a list of related terms. As each category corresponds to a WordNet synset, we use the lemmas for each synset to generate the related terms per concept. For categories with low frequency in the dataset, we supplement the list with related nouns (ie \textit{flute} when looking for examples of \textit{flutist}).
Separately, for each example (image, caption, tag), we tokenize and lemmatize the caption using the Natural Language Toolkit (NLTK) \cite{loper2002nltk}.
For instance, the caption ``The person is playing the flute" gets lemmatized to \{person, play, flute\} (without stop words).
We compute the overlap between the caption's lemmas + tags with the relevant term lists for each of the 52 categories to approximate which categories likely occur in each image.  We select images with the most overlap for annotation.

\subsection{Annotation Pipeline}

Given the sensitivity of these labels, we took multiple steps to ensure high-quality annotations.
Annotators completed stage-specific training before beginning labeling and \skintone annotations underwent multi-review.
Annotators could also mark \textit{cannot be determined} for any image where they could not perceive the attribute.
See Figure \ref{fig:pipeline} in Appendix \ref{app:benchmark_decisions}.

\paragraph{Stage 1:} \textit{Filtering Images by Target Categories}\label{stage1}
First, annotators are tasked with quickly filtering images based on whether they contain people who belong to a subset of categories.
Following the process described earlier in this section, we use the metadata
for each image to create a shorter list of likely categories per photo. 
We do not use any classification or detection models to filter images to avoid potential model biases that can skew the data distribution.
Annotators note the number of people in each image who match the specified categories. 
We exclude images marked with more than 5 people matching the target categories, given the time-intensive nature of annotating attributes for each person. This stage eliminates roughly 80\% of the candidate images.
For the remaining stages, we move from quick filtering to a focus on precision.

\paragraph{Stage 2:}\textit{ Annotating Bounding Boxes}\label{stage2}
Annotators are tasked with drawing bounding boxes around each person in the image that matches \textbf{any} of the target 52 categories.
For each bounding box, annotators mark a primary class, as well as a secondary class if necessary.
The primary and secondary class structure alleviates potential overlap between categories.
For example, a person playing the guitar and singing can match the category labels \texttt{guitarist} and \texttt{singer}.
Furthermore, allowing two classes permits for representation of visually ambiguous classes, \textit{e.g. a person in scrubs who could be a doctor or nurse}.

\paragraph{Stage 3:} \textit{Annotating \skintone}\label{stage3}
We assign \skintone annotations to its own step separate from other attributes to allow us to aggregate annotations from multiple raters. We choose to aggregate as one's own skin tone can affect the perceived skin tone of others \cite{10.2307/3088935, xdoi:10.1177/0002764215613401}.
In this stage, annotators label the \skintone of a specified person using the Monk Skin Tone Scale \cite{Monk_2019} (see Figure \ref{fig:mst}).
We ask annotators to select at least 2 adjacent values and aggregate the results across three annotators.
We report the number of times each skin tone was chosen.

\paragraph{Stage 4:} \textit{Annotating Remaining Attributes}\label{stage4}
In the final stage, annotators label the remaining attributes (see Section \ref{addn_attrs}) for each person in the bounding boxes from Stage 2.

\paragraph{Stage 5:} \textit{Annotating \datasourcenocite\ Masks}\label{mask_ann} As \datasetabbr images come from \datasourcelongnocite, which has images and masks, we label a subset of masks as \texttt{person, clothing, hair}. We do not collect exhaustive annotations for person-related masks in \datasetabbrnospace; we focus on annotating masks for people who are fully visible, with an attempt to balance demographic attributes. More details are given in Appendix \ref{app:sec:pipeline}.

\subsection{Annotator Diversity}

\begin{figure}
\resizebox {0.8\linewidth} {!} {
    \begin{tikzpicture}
        \pie{8.5/Egypt (14),
            24.2/Philippines (40),
            30.9/United States (51),
            12.1/Kenya (20),
            9.1/Spain (15),
            2.4/Taiwan (4),
            12.7/Colombia (21)}

    \end{tikzpicture}
    }
\caption{Breakdown of raters who passed training by percentage and by raw number, shown in parentheses.
}
\label{fig:annotator_pie1}
\end{figure}
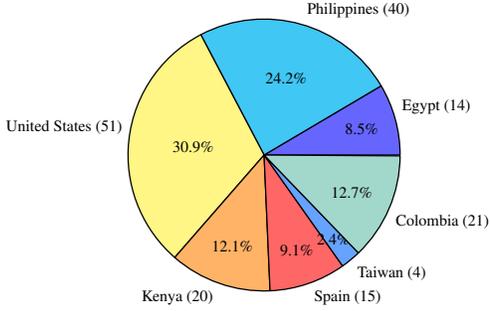

We prioritized having a geographically diverse set of annotators following \cite{kapania2023hunt, denton2021whose} and sourced raters from varying regions to increase the diversity of annotations. Our annotators come from six different geographic regions to increase the diversity of the annotations, with one country per region. These regions (with country in parenthesis) include North America (United States), Latin American (Colombia), Middle East (Egypt), Africa (Kenya), Southeast Asia (Philippines) and East Asia (Taiwan). We show more fine-grained breakdowns of annotators per region in Figure \ref{fig:annotator_pie1}. 
We aimed for a roughly balanced number of raters per region but had disproportionate pass-rates of training across the various regions. We further describe our annotation process and annotators sourced in Appendix \ref{app:ann_procedure}, and answer the questions posed by CrowdWorkSheets \cite{D_az_2022} in Appendix \ref{app:crowdworksheets}.

\subsection{FACET Statistics}

\begin{table}[!h]
\small{
  \centering
  \begin{tabular}{l|r r r r}

     \multicolumn{1}{c|}{\it \parbox{8em}{\centering Perceived or\\Apparent Attributes\\[1.2ex]}}
     & \multicolumn{1}{c|}{
     \rotatebox[origin=c]{90}{\ \it\#/people}
     }
     & \multicolumn{1}{c|}{\%}
     & \multicolumn{1}{c|}{
        \rotatebox[origin=c]{90}{\ \it\#/images}
        }
     & \multicolumn{1}{c|}{\parbox{2em}{\%}}  \\ \hline

gender presentation & & & &\\
\shortstack{\hspace{0.5em}-- more stereotypically F} & 10k & 21\% & 8k &  26\%  \\
\shortstack{\hspace{0.5em}-- more stereotypically M}   & 33k & 67\% & 23k &  72\%  \\
\shortstack{\hspace{0.5em}-- non-binary} & 95 & $<$1\% & 95 & $<$1\% \\
\shortstack{\hspace{0.5em}-- unknown}& 6k &      11\% &  5k & 5\% \\
\hline

Monk Skin Tone & & &  & \\
\hspace{0.5em} -- 1 \mstone &  5k & 10\% & 4k & 13\% \\\
\hspace{0.5em} -- 2  \msttwo  & 20k & 41\% & 15k & 48\%  \\
\hspace{0.5em} -- 3 \mstthree &  26k & 53\% & 19k & 61\% \\
\hspace{0.5em} -- 4 \mstfour  &  27k & 54\% & 20k & 63\% \\
\hspace{0.5em} -- 5 \mstfive  & 22k & 44\% & 17k &  54\% \\
\hspace{0.5em} -- 6 \mstsix  & 16k & 33\% & 13k & 40\% \\
\hspace{0.5em} -- 7 \mstseven  & 9k & 18\%  & 7k & 23\% \\
\hspace{0.5em} -- 8 \msteight  & 5k & 10\% & 4k & 13\% \\
\hspace{0.5em} -- 9 \mstnine  & 3k &  6\% & 2k &  7\% \\
\hspace{0.5em} -- 10 \mstten  & 1k & 3\% & 1k &   3\% \\
\hspace{0.5em} -- unknown & 18k & 37\%  & 13k &   42\%  \\
\hline

age & & & &\\
\hspace{0.5em}-- younger & 9k & 18\% &  7k &        23\%  \\
\hspace{0.5em}-- middle & 27k & 55\%  & 20k &        64\% \\
\hspace{0.5em}-- older & 3k & 5\%& 2k &         8\%  \\
\hspace{0.5em}-- unknown &  10k & 21\% &  9k &        27\% \\

\multicolumn{5}{c}{\parbox{25em}{\footnotesize \vspace*{2ex}*Images can have multiple labels for each attribute, which is why numbers may not sum to 100\%. F=femaleness; M=maleness}}
  \end{tabular}
}

\caption{Breakdown of representation of the demographic groups in the evaluation set.}
\label{tab:annotation_percentages}
\end{table}

\begin{figure}
\begin{center}
   \includegraphics[width=1.0\linewidth]{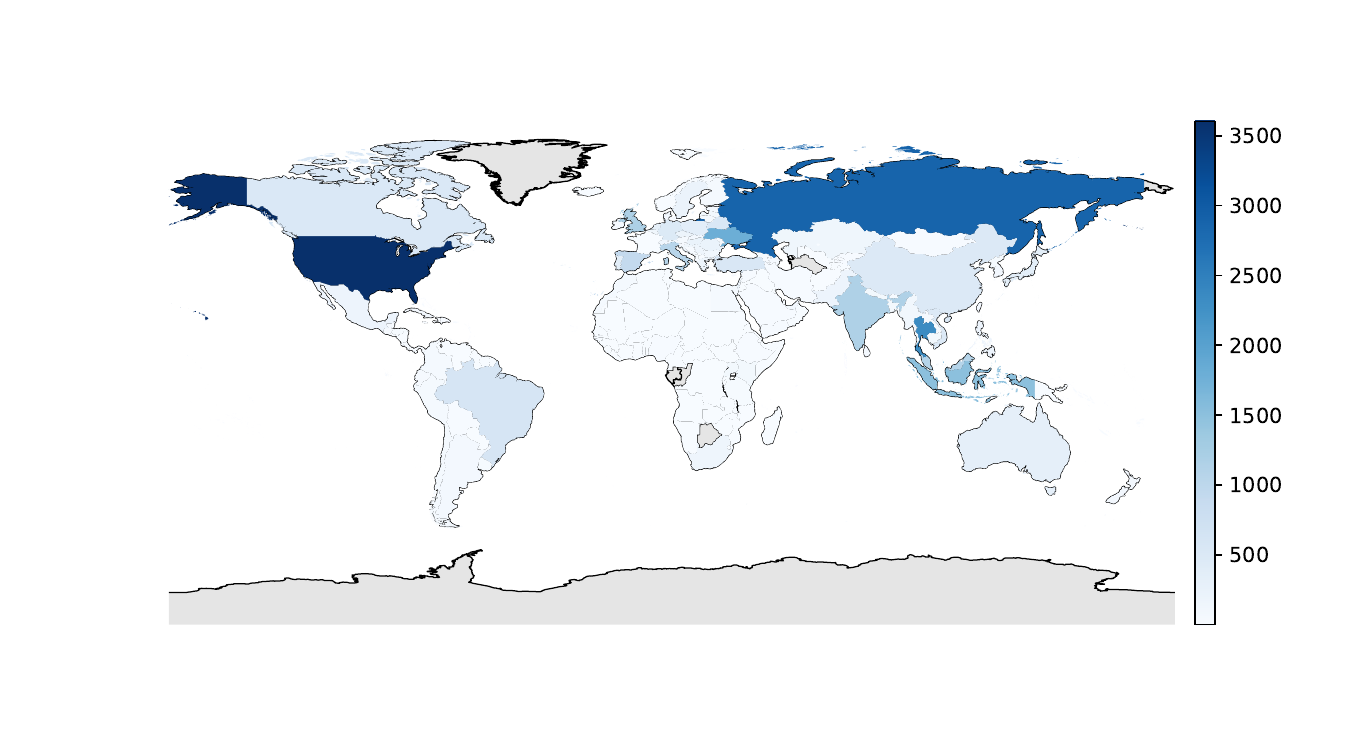}
\end{center}
\vspace*{-3em}
   \caption{Approximate geographic distribution of the images in FACET.}
\label{fig:facet_geo}
\end{figure}

In this section we summarize the attribute and image breakdown of FACET.
Table \ref{tab:annotation_percentages} 
shows the three demographic groups and their corresponding number of occurrences in the evaluation set.
The majority of \gender annotations are people perceived to have more stereotypically male attributes, followed by people perceived as having more stereotypically female attributes. \skintonecap annotations essentially follow a normal distribution; the majority of annotations are in the range of skin tones 3-6.
Appendix \ref{app:data_stats} details more statistics about the FACET benchmark including the number of people per class and demographic attribute along with the frequency of additional attributes. Figure \ref{fig:facet_geo} shows an approximate geographic breakdown of the images in FACET. The geographic information was inferred from locations mentioned in the captions, so the distribution is approximate.

\section{Fairness Evaluations using FACET}

We use \datasetabbr to evaluate fairness by measuring performance disparities across different attributes for a selection of state-of-the-art vision models.
Given a model $f$, a performance metric \textit{recall}, a set of concepts $\mathcal{C}$, an attribute label \textit{l} and a set of images $\mathcal{I}_{l}^{C}$, 
we compute:\vspace*{-.5ex}

\begin{equation}
\begin{split}
disparity =\  & recall( f(l_{1}, \mathcal{I}_{l_{1}}^{C}, \mathcal{C}))  \\
    & \ \ \ \ \ \ \ \ \ \quad
    - recall(f(l_{2}, \mathcal{I}_{l_{2}}^{C}, \mathcal{C}))
\end{split}
\end{equation}

As a concrete example, we can compute the disparity between people perceived as younger  ($l_{1}$) versus people perceived as older ($l_{2}$) for the concept teacher ($\mathcal{C}=\{\textrm{teacher}\}$).
Images $\mathcal{I}_{l_{1}}^{C}$ and $\mathcal{I}_{l_{2}}^{C}$ are images of teachers who are perceived to be younger and images of teachers who are perceived to be older, respectively.
Disparity $>$ 1 indicates the model performs better for images with label $l_{1}$ and disparity $<$ 1 indicates the model performs  better for label $l_{2}$.
A perfectly fair model evaluated with \datasetabbr would have a disparity of 0, meaning it has the same performance across all images regardless of the associated attributes.

\datasetabbr is unique for two key reasons:
\begin{enumerate}[noitemsep,nolistsep]
    \item[1.] \textit{Exhaustive attribute and class level evaluation:} \datasetabbrnospace's annotations are exhaustive, meaning every person who matches a class in every image is annotated across all attributes through a rigorous annotation process.
    Datasets that include only the \texttt{person} class label and/or very sparse attribute labels risk bias leakage from the unlabeled people or difficulty performing a deep analysis due to the lack of exhaustive labels. These are not concerns with \datasetabbrnospace.
    \item[2.] \textit{Annotations for multiple vision tasks:} Because every image is annotated with bounding boxes and person-related classes, multiple vision tasks can be evaluated and analyzed alongside the exhaustively annotated person attributes.
\end{enumerate}

\subsection{Classification}
\notab\textbf{Are models better at classifying people as skateboarder when their \gender has more stereotypically male attributes?}
To help answer this question and others like it, we evaluate standard image classification models with the FACET class labels.
For classification evaluation, we only evaluate images with a single person.
For images where a single person is labeled with multiple person classes, we treat both classes as valid labels.
We evaluate classification using CLIP ViT-B/32 \cite{radford2021learning} in a zero-shot setting.
The largest discrepancies for CLIP on the \gender axis are shown in Table \ref{fig:clip}.
Some of these classes parallel societal, gender-related biases (\eg higher performance on careers like nurses and hairdressers for those who are perceived with more femaleness \cite{Greenwald1998MeasuringID,nosek2002harvesting}).
We show further analysis of CLIP across other demographic groups in Appendix \ref{app:evaluation}.
We also show how we can use the FACET IN21k class overlap to evaluate an ImageNet21k pre-trained ViT \cite{dosovitskiy2020image}.

\begin{table}
\centering
{\it Difference in Average Recalls}\\

\begin{tabular}{l|r|}

Person Class &  {\lightBlue${M}$} $-$ {\orange $F$} \\
\hline
gardener     &         16.4 \\
craftsman    &         13.6 \\
laborer      &         10.3 \\
skateboarder &          8.8 \\
prayer       &          8.8 \\
waiter       &          8.3 \\
speaker      &          5.4 \\
guitarist    &          4.0 \\
singer       &          1.6 \\
lawman       &          1.4 \\
\end{tabular}
\begin{tabular}{|l|r}

Person Class &  {\orange${F}$} $-$ {\lightBlue $M$} \\
\hline
dancer      &        21.7 \\
retailer    &        17.0 \\
reporter    &        16.0 \\
nurse       &        12.9 \\
student     &        12.8 \\
gymnast     &         8.5 \\
painter     &         6.1 \\
hairdresser &         5.2 \\
climber     &         5.1 \\
horseman    &         4.5 \\
\end{tabular}
\caption{CLIP's performance disparity for the classes with the largest disparity across \gendernospace. The classes on the left indicate better performance for images with people who are perceived as having {\textbf{\lightBlue more stereotypically male attributes}}; results on the right indicate better performance for those perceived as having {\textbf{\orange more stereotypically female attributes}}.}
\label{fig:clip}
\end{table}

\subsection{Person Detection \& Segmentation}
\subsubsection{Person Detection}

We evaluate a Faster R-CNN model with a ResNet-50-FPN backbone \cite{ren2015faster} pretrained on COCO.
During evaluation, we only keep the predicted boxes corresponding to the COCO \texttt{person} class.
We treat the remaining boxes as class-agnostic, and we compute the average recall (AR) and mAR (mean average recall) metrics proposed in \cite{DBLP:journals/corr/HosangBDS15} with all predicted boxes
and measure
performance across the demographic attributes.
For person detection, we focus on evaluating \skintone and how model performance parallels societal biases \cite{dixon2005skin,kleider2017black}. 

\paragraph{Do standard detection models struggle to detect people whose skin appears darker?}
We compute the AR across the predicted bounding boxes for \skintonenospace, as shown in Table \ref{tab:maskrcnn}.
At \textit{every} IoU for the ARs, the darkest \skintone has the lowest performance.
The gap between the highest and lowest performance is over 4 points at AR$_{0.75}$, which suggests Faster R-CNN does struggle more on precisely detecting those perceived with darker skin tones.

\paragraph{Does this problem magnify when, for instance, the person has coily compared to straight hair?}
We dig deeper into the previous results to investigate intersections of attributes. In Table \ref{table:hairtype} we measure mAR per hair type for people with the three lightest \skintone versus the three darkest \skintonenospace.
This unearths several concerning findings: 
for 456 hair types performance is higher for the lighter skin tones than the darker skin tones.
These are fairly significant gaps; for instance, we see a nearly 10 point difference for hair type \textit{dreads} at AR$_{0.75}$.
This is a particularly interesting finding for two reasons.
First, we see a 50x increase in the disparity for \textit{dreads} across \skintone from AR$_{0.5}$ to AR$_{0.75}$.
This suggests Faster R-CNN can detect people with dreadlocks, but struggles to perform accurate localization as shown by the larger gap as the IoU threshold for AR increases. 
Second, dreadlocks are often associated with darker skin and a plethora of associated stereotypes \cite{donahoo2022controlling,byrd2014black,prendergast2016case,banks2021no}).
This means the likely association between \textit{dreads} and darker skin tones in the training data interestingly combat this performance disparity.

\begin{table}
\begin{center}
\begin{tabular}{l|c c c}
\it Monk Skin Tone (MST) & mAR & \multicolumn{1}{|c|}{AR$_{0.5}$} & \multicolumn{1}{c|}{AR$_{0.75}$} \\\hline

\hspace{1.5em}-- 1 \mstone  &  75.5  & 	98.4  & 	85.0 \\
\hspace{1.5em}-- 2 \msttwo  & 75.0  & 	98.3  & 	84.0 \\
\hspace{1.5em}-- 3  \mstthree & 74.7  & 	98.3  & 	83.5 \\
\hspace{1.5em}-- 4 \mstfour  &  74.4  & 	98.1  & 	83.0 \\
\hspace{1.5em}-- 5 \mstfive  & 74.1  & 	98.2  & 	82.6 \\
\hspace{1.5em}-- 6 \mstsix  &	73.9  & 	98.3  & 	82.5 \\
\hspace{1.5em}-- 7 \mstseven & 73.7  & 	98.2  & 	82.2 \\
\hspace{1.5em}-- 8 \msteight & 73.7  & 	98.0  & 	82.5 \\
\hspace{1.5em}-- 9 \mstnine  & 73.3  & 	97.3  & 	81.1 \\
\hspace{1.5em}-- 10 \mstten  & 72.6 & 96.5 & 80.4  \\

\hline
\multicolumn{4}{c}{
    \parbox{20em}{
    \footnotesize\vspace*{1.5ex} Average Recall (AR), with IoU values as subscripts. mAR is averaged across IoUs from 0.5 to 0.95, in increments of 0.05.
    }
}
\end{tabular}
\caption{Average recall (AR) on FACET for a ResNet-50 Faster R-CNN. The model has the best performance for MST=1, which is the lightest skin tone, and the lowest performance for MST=9 and 10, which are the darkest skin tones. The largest disparity between skin tones is for AR$_{0.75}$.
} 
\label{tab:maskrcnn}
\end{center}
\end{table}

\begin{table}

\begin{center}
    \textit{lighter} = $\{$1 {\mstone}, 2 {\msttwo}, 3 {\mstthree} $\}$
    \hspace{0.5em}\textit{darker} = $\{$8 {\msteight}, 9 {\mstnine}, 10 {\mstten} $\}$\\[2ex]
    
    {
    \begin{tabular}{l | c c | c c | c c}
    \multirow{2}{*}{
    \parbox{2.6em}{\centering\small\textit{Hair {\footnotesize $\downarrow$}\\Skin {\scriptsize$\rightarrow$}}}
    }
    & \multicolumn{2}{c|}{mAR}
    & \multicolumn{2}{c|}{AR$_{0.5}$}
    & \multicolumn{2}{c|}{AR$_{0.75}$}
    \\
    
    & \multicolumn{1}{c|}{\footnotesize lighter}
    & \multicolumn{1}{c|}{\footnotesize darker}
    & \multicolumn{1}{c|}{\footnotesize lighter}
    & \multicolumn{1}{c|}{\footnotesize darker}
    & \multicolumn{1}{c|}{\footnotesize lighter}
    & \multicolumn{1}{c|}{\footnotesize darker} \\\hline

    \small coily & 
    \bf 76.7   & 73.4 & 
     98.2 & \bf 98.5 &
    \bf 87.3 & 80.9  \\
    
    \small dreads & 
    \bf 77.1  & 74.7   & 
     97.9 & \bf 98.1 &
    \bf 94.8 & 85.7  \\
    
    \small bald & 
    \bf 78.1 & 71.5 & 
    \bf 99.0 &  96.7 & 
    \bf 87.8  & 77.5 \\
    
    \small straight & 
    75.6 & \bf 76.1  & 
    98.4 & \bf  99.1 &
    84.8 & \bf 85.6  \\
    
    \small curly & 
    \bf 75.0 & 74.8 &
    98.5 & \bf 99.2 &
    \bf 84.7 & 83.7 \\
    
    \small wavy & 
    \bf 76.1 & 75.8  &
    98.6 & \bf  99.1  & 
    \bf 85.5 & 84.8 \\[1.3ex]

    \multicolumn{7}{c}{
    \parbox{21em}{
    \footnotesize\vspace*{1.5ex} Average Recall (AR), with IoU values as subscripts. mAR is averaged across IoUs from 0.5 to 0.95, in increments of 0.05.
    }
    }
    
    \end{tabular}
}
\end{center}
\vspace{-0.5em}
\caption{
Average recall (AR) on FACET for a ResNet-50 Faster R-CNN. We show performance for the intersection of hair type and \skintonenospace.
Performance is higher for lighter \skintone  for every hair type except straight and wavy.
}
\label{table:hairtype}
\end{table}

\subsubsection{Person Instance Segmentation}

We evaluate a Mask R-CNN model with a ResNet-50-FPN backbone \cite{he2017mask} pretrained on COCO.
In the same pattern as for person detection, we only keep the predicted masks corresponding to the COCO \texttt{person} class, and compute AR in a class agnostic way. We use the IoU between predicted and ground truth masks for instance segmentation, opposed to between boxes used for detection.

\paragraph{Do performance discrepancies differ across the
detection and segmentation task? } We compare potential discrepancies across segmentation and detection of people.
We evaluate Mask R-CNN for person detection and person segmentation separately. For consistency, we limit the evaluation for both detection and segmentation to the set of people who have a mask annotation.  We compare the patterns of discrepancies in AR across \gender for person detection and segmentation, as shown in Table \ref{tab:maskrcnn_seg_people}.  We notice that for both detection and segmentation, the performance disparities are largest at AR$_{0.75}$. We also observe slightly larger gaps in performance for detection compared to segmentation. In line with prior work \cite{he2017mask}, we find higher AR for person detection than instance segmentation.
We describe the experimental setup in more detail in Appendix \ref{app:evaluation}.

\renewcommand{\tabcolsep}{3pt}
\begin{table}[h]
\centering\small
\begin{tabular}{l|c c | c c | c c|}
& \multicolumn{2}{c|}{\normalsize mAR}
& \multicolumn{2}{c|}{\normalsize AR$_{0.5}$}
& \multicolumn{2}{c|}{\normalsize AR$_{0.75}$}\\

\scalebox{1.1}{\parbox{8em}{\centering\it perceived gender \\ presentation\vspace*{0.5ex}}}
& \multicolumn{1}{c|}{\normalsize \rotatebox{90}{\parbox{2.8em}{$box$}}}
& \multicolumn{1}{c|}{\normalsize \rotatebox{90}{\parbox{2.8em}{$mask$}}}
& \multicolumn{1}{c|}{\normalsize \rotatebox{90}{\parbox{2.8em}{$box$ }}}
& \multicolumn{1}{c|}{\normalsize \rotatebox{90}{\parbox{2.8em}{$mask$ }}}
& \multicolumn{1}{c|}{\normalsize \rotatebox{90}{\parbox{2.8em}{$box$ }}}
& \multicolumn{1}{c|}{\normalsize \rotatebox{90}{\parbox{2.8em}{$mask$ }}}\\ \hline

\scalebox{1}{\parbox{9em}{\vspace*{0.5ex}-- more stereotypically \\\hspace*{0.8em}male attributes\vspace*{0.8ex}}}      & 	78.3 & 72.2 &    	99.3 &98.1 &   	88.0 &84.6  \\ \hline

\scalebox{1}{\parbox{9em}{\vspace*{0.5ex}-- more stereotypically\\\hspace*{0.5em} female attribute\vspace*{0.8ex}s }}        & 	75.6 & 70.8 &    	99.0 &97.5 &   	84.7 &82.9  \\\hline

\scalebox{1}{\parbox{9em}{\vspace*{0.5ex}-- outside of gender\\\hspace*{0.8em}binary}} & 	77.0 & 63.0 &    	98.0 &92.0 &   	88.0 &74.0  \\[1.3ex]

\multicolumn{7}{l}{
    \centering
    \hspace*{1em}\parbox{23em}{
    \footnotesize\vspace*{2ex} Average Recall (AR), with IoU values as subscripts. mAR is averaged across IoUs from 0.5 to 0.95, in increments of 0.05.
    }
}
\end{tabular}
\caption{We compare the AR on FACET for a ResNet-50 Mask R-CNN across the person detection and person instance segmentation tasks. The candidates $box$ dictates the AR for person detection, box proposals, and $mask$ for segmentation, mask proposals.
}
\label{tab:maskrcnn_seg_people}
\end{table}
\subsection{Open World Detection \& Visual Grounding}
\subsubsection{Open Vocabulary Detection}
Next we evaluate open vocabulary detection using Detic \cite{zhou2022detecting}.
We describe the experimental setup in detail in the Appendix \ref{app:evaluation}.
For Detic, we focus on \agenospace.

\paragraph{Are open-vocabulary detection models better at detecting backpackers who are perceived to be younger?}
To be illustrative of disparities observed with FACET, we selected the three person-related classes with the biggest disparity between groups:\textit{ trumpeter, backpacker and painter}. Detic exhibits \agenospace-based performance disparities for all 3 categories. The disparities are large, with a 15 point gap in mAR for the \textit{backpacker} class.
The disparities are also consistent across AR measurements for a specific class.
Unlike what we observed for Faster R-CNN with \skintonenospace, we typically observe larger gaps for AR$_{0.5}$ than AR$_{0.75}$.
This suggest that in open-world detection, there is perhaps more of a discrepancy in the classification/identifying a person for a given category than there is in the precision of the bounding box.
We show all per-class disparities as well as the  mean disparities across all \numpersonclasses classes in Table \ref{tab:age_both} in Appendix \ref{app:evaluation}.

\subsubsection{Visual Grounding}

Lastly, we evaluate visual grounding using OFA \cite{wang2022unifying}, a sequence-to-sequence vision-language model.
We evaluate \age disparities using three person classes with large disparities which are nurse, gardener and guitarist.
Results are in Table \ref{tab:age_both}.
OFA's largest disparity is nearly 27 points, observed in the nurse class.
The disparities and best performing \age label for every class is consistent across all IoU values.
We show the full table of per-class disparities as well as the disparities averaged across all classes in Table \ref{tab:age_both} in Appendix \ref{app:evaluation}.

\renewcommand{\tabcolsep}{4pt}
\begin{table}[h]
\centering\small
\begin{tabular}{l|c c c || c c c}
\it \multirow{2}{*}{\parbox{0.8em}{\centering\small Person\\ Class}}
& \multicolumn{3}{c||}{Detic (detection)}
& \multicolumn{3}{c}{OFA (visual grounding)}\\

& \multicolumn{1}{c|}{\footnotesize mAR}
& \multicolumn{1}{c|}{\footnotesize AR$_{0.5}$}
& \multicolumn{1}{c||}{\footnotesize AR$_{0.75}$} 
& \multicolumn{1}{c|}{\footnotesize mAR}
& \multicolumn{1}{c|}{\footnotesize AR$_{0.5}$}
& \multicolumn{1}{c}{\footnotesize AR$_{0.75}$} \\\hline

& \multicolumn{3}{c||}{\it backpacker}
& \multicolumn{3}{c}{\it gardener}\\
\small -- young
&  \bf 45.4 & \bf 55.3 & \bf 47.4 
& 11.9 & 32.6 & 7.0 \\

\small -- middle
& 42.1 & 51.7 & 44.6 
& 18.3 & 40.1 & 14.6 \\

\small -- older &
29.8 & 35.3 & 33.3 
& \bf 27.9 & \bf 58.1 & \bf 24.2 \\
\hline

& \multicolumn{3}{c||}{\it trumpeter}
& \multicolumn{3}{c}{\it solider}\\
\small -- young
& 22.8 & 26.7 & 25.6
& \bf 16.3 & 9.2 & \bf 40.0  \\
\small -- middle
&  29.5 & 34.8 & 31.4 
& \textbf{16.3} & \textbf{14.5} & 33.8 \\
\small -- older
& \bf 38.4 & \bf 45.5 &\bf 38.2 
& 1.3 & 6.3 & 0.0 \\
\hline

& \multicolumn{3}{c||}{\it drummer}
& \multicolumn{3}{c}{\it guitarist}\\
\small -- young
& 19.9 & 24.9 & 19.7
& 19.8 & 38.7 & 18.5 \\
\small -- middle
&  26.3 & 34.7 & 27.6 
& 19.6 & 35.5 & 20.6 \\
\small -- older
& \bf 34.2 & \bf 41.8 & \bf 35.8
& \bf32.0 & \bf56.7 & \bf36.7 \\[1.2ex]

\multicolumn{7}{c}{
    \parbox{22em}{
    \footnotesize\vspace*{0.2ex} Average Recall (AR), with IoU values as subscripts. mAR is averaged across IoUs from 0.5 to 0.95, in increments of 0.05.
    }
}
\end{tabular}
\caption{Per-class performance for Detic and OFA on a subset of \datasetabbr classes. The \age with the highest performance
per class is bolded.
}
\label{fig:ofa_age_per_class}
\end{table}

\subsection{Limitations}\label{sec:limitations}
As the development of datasets for fairness analysis becomes more common, approaches and recommendations for how to do so in ethical and safe ways are being increasingly explored \cite{ccv2,gebru2021datasheets,andrews2023ethical,10.1145/3392866}. 
While we strongly believe FACET will help practitioners better understand sources of bias in their model, we note that translating real world concepts and demographic groups to dataset annotations is inherently imperfect.

First, while self-identification of concept classes and person-related attributes is preferred \cite{10.1145/3392866,https://doi.org/10.48550/arxiv.2210.11924}, our adaptation of an existing dataset requires external annotations.
To reduce these potential biases, we use highly trained annotators and avoid automated labeling methods like adapting existing captions, alt-text or model classifications \cite{zhao2017men,DirectionalBiasAmp,https://doi.org/10.48550/arxiv.2210.11924,hall2023vision,hall2023towards,MIAP,buolamwini2018gender}.
Second,
while generating sets of labels for each attribute, there is a trade-off between having more labels (wider representation) and opting for fewer, higher frequency labels (more statistical significance) \cite{hall2023towards}.
This was extensively considered, and we acknowledge that, as with any paper using discrete labels, our labels for \gender and \age risk erasure of genders and ages that are not identifiable in our categorization \cite{https://doi.org/10.48550/arxiv.2205.02526}.
For skin tone, we follow the Monk Scale \cite{Monk_2019}, which shows better inclusivity of darker skin tones.  
For concept classes, we map to ImageNet classes to encourage easy adoption and to ensure mutually exclusive classes.
Lastly, FACET and other fairness datasets are representative of the current time period and organizational infrastructure within which it was created \cite{doi:10.1177/20539517211035955,gebru2021datasheets}.
To address how this affects annotations and insights when performing evaluations, we include in Appendix \ref{app:crowdworksheets} our responses to the CrowdWorkSheets \cite{D_az_2022} for FACET.
\section{Discussion}

We have seen rapid growth and impressive performance gains in computer vision across a number of tasks such as classification, detection, segmentation and visual grounding.
Simultaneously, these models have learned societal biases and can perpetuate these harmful stereotypes in downstream tasks.
We present \datasetabbrnospace, a vision fairness benchmark that contains \datasetsizemini annotated images of \peoplesizemini people.
People in the images are exhaustively labeled with demographic attributes, including \gendernospace, \skintone and \agenospace, and additional attributes such as hair type and light exposure. Labeling demographic attributes requires thoughtful design, so we hired expert annotators and prioritized clean annotations.
In addition to these attributes, \datasetabbr also has manual annotations for bounding boxes and person-related classes. 
These person-related classes, such as hairdresser and farmer, overlap with the ImageNet-21K (IN21k) vocabulary, meaning vision models that can be evaluated on IN21k can also seamlessly use \datasetabbrnospace.
We aimed to be extremely conscious and respectful with our annotations, while also acknowledging that there are limitations with this and similar fairness datasets.
We are publicly releasing \datasetabbr to encourage and lower the barrier to entry to evaluating vision models for potential biases. We propose several ways that researchers can use FACET to evaluate their models for potential fairness concerns across a variety of common vision tasks.
\paragraph{Acknowledgments}
We would like to thank Adina Williams, Priya Goyal, Zachary Winstrom, Austin Miller, Nicolas Carion,  Nisha Deo, Emily Dinan, Ahuva Goldstand,  Somya Jain, Alexander Kirillov, Mallika Malhotra, Hanzi Mao, Ishan Misra, Laurens van der Maaten, and Mary Williamson.

{\small
\bibliographystyle{ieee_fullname}
\bibliography{egbib}

\begin{thebibliography}{100}\itemsep=-1pt

\bibitem{dollarstreet}
{Dollar Street}.
\newblock \url{https://www.gapminder.org/dollar-street}.
\newblock Accessed: 2023-03-08.

\bibitem{UN}
{Provisional Guidelines on Standard International Age Classifications}.
\newblock
  \url{https://unstats.un.org/unsd/publication/SeriesM/SeriesM_74e.pdf}.
\newblock Accessed: 2023-03-08.

\bibitem{andrews2023ethical}
Jerone~TA Andrews, Dora Zhao, William Thong, Apostolos Modas, Orestis
  Papakyriakopoulos, Shruti Nagpal, and Alice Xiang.
\newblock Ethical considerations for collecting human-centric image datasets.
\newblock {\em arXiv preprint arXiv:2302.03629}, 2023.

\bibitem{banks2021no}
Patricia~A Banks.
\newblock No dreadlocks allowed: Race, hairstyles, and cultural exclusion in
  schools.
\newblock {\em Hairstyles, and Cultural Exclusion in Schools (February 12,
  2021)}, 2021.

\bibitem{barbu2019objectnet}
Andrei Barbu, David Mayo, Julian Alverio, William Luo, Christopher Wang, Dan
  Gutfreund, Josh Tenenbaum, and Boris Katz.
\newblock Objectnet: A large-scale bias-controlled dataset for pushing the
  limits of object recognition models.
\newblock {\em Advances in neural information processing systems}, 32, 2019.

\bibitem{birhane2021multimodal}
Abeba Birhane, Vinay~Uday Prabhu, and Emmanuel Kahembwe.
\newblock Multimodal datasets: misogyny, pornography, and malignant
  stereotypes.
\newblock {\em arXiv preprint arXiv:2110.01963}, 2021.

\bibitem{brendel2019approximating}
Wieland Brendel and Matthias Bethge.
\newblock Approximating cnns with bag-of-local-features models works
  surprisingly well on imagenet.
\newblock {\em arXiv preprint arXiv:1904.00760}, 2019.

\bibitem{buolamwini2018gender}
Joy Buolamwini and Timnit Gebru.
\newblock Gender shades: Intersectional accuracy disparities in commercial
  gender classification.
\newblock In {\em Conference on fairness, accountability and transparency},
  pages 77--91. PMLR, 2018.

\bibitem{byrd2014black}
Ayana Byrd and Lori~L Tharps.
\newblock When black hair is against the rules.
\newblock {\em The New York Times}, 1, 2014.

\bibitem{doi:10.1177/000312240707200505}
Mary~E. Campbell and Lisa Troyer.
\newblock The implications of racial misclassification by observers.
\newblock {\em American Sociological Review}, 72(5):750--765, 2007.

\bibitem{carion2020end}
Nicolas Carion, Francisco Massa, Gabriel Synnaeve, Nicolas Usunier, Alexander
  Kirillov, and Sergey Zagoruyko.
\newblock End-to-end object detection with transformers.
\newblock In {\em European conference on computer vision}, pages 213--229.
  Springer, 2020.

\bibitem{crawford_paglen_2019}
Kate Crawford and Trevor Paglen.
\newblock Excavating ai: The politics of training sets for machine learning.
\newblock \url{https://excavating.ai/}, September 2019.

\bibitem{crenshaw1989demarginalizing}
Kimberl{\'e} Crenshaw.
\newblock Demarginalizing the intersection of race and sex: A black feminist
  critique of antidiscrimination doctrine, feminist theory and antiracist
  politics.
\newblock {\em u. Chi. Legal f.}, page 139, 1989.

\bibitem{daniel2018quality}
Florian Daniel, Pavel Kucherbaev, Cinzia Cappiello, Boualem Benatallah, and
  Mohammad Allahbakhsh.
\newblock Quality control in crowdsourcing: A survey of quality attributes,
  assessment techniques, and assurance actions.
\newblock {\em ACM Computing Surveys (CSUR)}, 51(1):1--40, 2018.

\bibitem{de2019does}
Terrance De~Vries, Ishan Misra, Changhan Wang, and Laurens Van~der Maaten.
\newblock Does object recognition work for everyone?
\newblock In {\em Proceedings of the IEEE/CVF Conference on Computer Vision and
  Pattern Recognition Workshops}, pages 52--59, 2019.

\bibitem{imagenet_cvpr09}
J. Deng, W. Dong, R. Socher, L.-J. Li, K. Li, and L. Fei-Fei.
\newblock {ImageNet: A Large-Scale Hierarchical Image Database}.
\newblock In {\em CVPR09}, 2009.

\bibitem{denton2021whose}
Emily Denton, Mark D{\'\i}az, Ian Kivlichan, Vinodkumar Prabhakaran, and Rachel
  Rosen.
\newblock Whose ground truth? accounting for individual and collective
  identities underlying dataset annotation.
\newblock {\em arXiv preprint arXiv:2112.04554}, 2021.

\bibitem{doi:10.1177/20539517211035955}
Emily Denton, Alex Hanna, Razvan Amironesei, Andrew Smart, and Hilary Nicole.
\newblock On the genealogy of machine learning datasets: A critical history of
  imagenet.
\newblock {\em Big Data \& Society}, 8(2):20539517211035955, 2021.

\bibitem{denton2020bringing}
Emily Denton, Alex Hanna, Razvan Amironesei, Andrew Smart, Hilary Nicole, and
  Morgan~Klaus Scheuerman.
\newblock Bringing the people back in: Contesting benchmark machine learning
  datasets.
\newblock {\em arXiv preprint arXiv:2007.07399}, 2020.

\bibitem{https://doi.org/10.48550/arxiv.2205.02526}
Hannah Devinney, Jenny Björklund, and Henrik Björklund.
\newblock Theories of "gender" in nlp bias research, 2022.

\bibitem{D_az_2022}
Mark D{\'{\i}}az, Ian Kivlichan, Rachel Rosen, Dylan Baker, Razvan Amironesei,
  Vinodkumar Prabhakaran, and Emily Denton.
\newblock {CrowdWorkSheets}: Accounting for individual and collective
  identities underlying crowdsourced dataset annotation.
\newblock In {\em 2022 {ACM} Conference on Fairness, Accountability, and
  Transparency}. {ACM}, jun 2022.

\bibitem{dixon2005skin}
Travis~L Dixon and Keith~B Maddox.
\newblock Skin tone, crime news, and social reality judgments: Priming the
  stereotype of the dark and dangerous black criminal 1.
\newblock {\em Journal of Applied Social Psychology}, 35(8):1555--1570, 2005.

\bibitem{DBLP:journals/corr/abs-2006-07397}
Brian Dolhansky, Joanna Bitton, Ben Pflaum, Jikuo Lu, Russ Howes, Menglin Wang,
  and Cristian Canton{-}Ferrer.
\newblock The deepfake detection challenge dataset.
\newblock {\em CoRR}, abs/2006.07397, 2020.

\bibitem{donahoo2022controlling}
Saran Donahoo and Asia~D Smith.
\newblock Controlling the crown: Legal efforts to professionalize black hair.
\newblock {\em Race and Justice}, 12(1):182--203, 2022.

\bibitem{dooley2021comparing}
Samuel Dooley, Ryan Downing, George Wei, Nathan Shankar, Bradon Thymes, Gudrun
  Thorkelsdottir, Tiye Kurtz-Miott, Rachel Mattson, Olufemi Obiwumi, Valeriia
  Cherepanova, et~al.
\newblock Comparing human and machine bias in face recognition.
\newblock {\em arXiv preprint arXiv:2110.08396}, 2021.

\bibitem{dosovitskiy2020image}
Alexey Dosovitskiy, Lucas Beyer, Alexander Kolesnikov, Dirk Weissenborn,
  Xiaohua Zhai, Thomas Unterthiner, Mostafa Dehghani, Matthias Minderer, Georg
  Heigold, Sylvain Gelly, et~al.
\newblock An image is worth 16x16 words: Transformers for image recognition at
  scale.
\newblock {\em arXiv preprint arXiv:2010.11929}, 2020.

\bibitem{self-report-fitzpatrick}
Steven Eilers, Daniel Bach, Rikki Gaber, Hanz Blatt, Yanina Guevara, Katie
  Nitsche, Roopal Kundu, and June Robinson.
\newblock Accuracy of self-report in assessing fitzpatrick skin phototypes i
  through vi.
\newblock {\em JAMA dermatology (Chicago, Ill.)}, 149, 09 2013.

\bibitem{xdoi:10.1177/0002764215613401}
Cynthia Feliciano.
\newblock Shades of race: How phenotype and observer characteristics shape
  racial classification.
\newblock {\em American Behavioral Scientist}, 60(4):390--419, 2016.

\bibitem{fitzpatrick1975soleil}
Thomas~B Fitzpatrick.
\newblock Soleil et peau.
\newblock {\em J. Med. Esthet.}, 2:33--34, 1975.

\bibitem{gatys2015texture}
Leon Gatys, Alexander~S Ecker, and Matthias Bethge.
\newblock Texture synthesis using convolutional neural networks.
\newblock {\em Advances in neural information processing systems}, 28, 2015.

\bibitem{gatys2017texture}
Leon~A Gatys, Alexander~S Ecker, and Matthias Bethge.
\newblock Texture and art with deep neural networks.
\newblock {\em Current opinion in neurobiology}, 46:178--186, 2017.

\bibitem{gebru2021datasheets}
Timnit Gebru, Jamie Morgenstern, Briana Vecchione, Jennifer~Wortman Vaughan,
  Hanna Wallach, Hal~Daum{\'e} Iii, and Kate Crawford.
\newblock Datasheets for datasets.
\newblock {\em Communications of the ACM}, 64(12):86--92, 2021.

\bibitem{geirhos2018imagenettrained}
Robert Geirhos, Patricia Rubisch, Claudio Michaelis, Matthias Bethge, Felix~A.
  Wichmann, and Wieland Brendel.
\newblock Imagenet-trained {CNN}s are biased towards texture; increasing shape
  bias improves accuracy and robustness.
\newblock In {\em International Conference on Learning Representations}, 2019.

\bibitem{goyal2022fairness}
Priya Goyal, Adriana~Romero Soriano, Caner Hazirbas, Levent Sagun, and Nicolas
  Usunier.
\newblock Fairness indicators for systematic assessments of visual feature
  extractors.
\newblock {\em arXiv preprint arXiv:2202.07603}, 2022.

\bibitem{Greenwald1998MeasuringID}
Anthony~G Greenwald, Debbie~E. McGhee, and Jordan L.~K. Schwartz.
\newblock Measuring individual differences in implicit cognition: the implicit
  association test.
\newblock {\em Journal of personality and social psychology}, 74 6:1464--80,
  1998.

\bibitem{10.1145/3555634}
Matthew Groh, Caleb Harris, Roxana Daneshjou, Omar Badri, and Arash Koochek.
\newblock Towards transparency in dermatology image datasets with skin tone
  annotations by experts, crowds, and an algorithm.
\newblock {\em Proc. ACM Hum.-Comput. Interact.}, 6(CSCW2), nov 2022.

\bibitem{gustafson2023pinpointing}
Laura Gustafson, Megan Richards, Melissa Hall, Caner Hazirbas, Diane
  Bouchacourt, and Mark Ibrahim.
\newblock Pinpointing why object recognition performance degrades across income
  levels and geographies.
\newblock {\em arXiv preprint arXiv:2304.05391}, 2023.

\bibitem{hall2023towards}
Melissa Hall, Bobbie Chern, Laura Gustafson, Denisse Ventura, Harshad Kulkarni,
  Candace Ross, and Nicolas Usunier.
\newblock Towards reliable assessments of demographic disparities in
  multi-label image classifiers.
\newblock {\em arXiv preprint arXiv:2302.08572}, 2023.

\bibitem{hall2023vision}
Melissa Hall, Laura Gustafson, Aaron Adcock, Ishan Misra, and Candace Ross.
\newblock Vision-language models performing zero-shot tasks exhibit
  gender-based disparities.
\newblock {\em arXiv preprint arXiv:2301.11100}, 2023.

\bibitem{10.1145/3173574.3173582}
Foad Hamidi, Morgan~Klaus Scheuerman, and Stacy~M. Branham.
\newblock Gender recognition or gender reductionism? the social implications of
  embedded gender recognition systems.
\newblock In {\em Proceedings of the 2018 CHI Conference on Human Factors in
  Computing Systems}, CHI '18, page 1–13, New York, NY, USA, 2018.
  Association for Computing Machinery.

\bibitem{hardt2016equality}
Moritz Hardt, Eric Price, and Nati Srebro.
\newblock Equality of opportunity in supervised learning.
\newblock {\em Advances in neural information processing systems}, 29, 2016.

\bibitem{ccv2}
Caner Hazirbas, Yejin Bang, Tiezheng Yu, Parisa Assar, Bilal Porgali, Vítor
  Albiero, Stefan Hermanek, Jacqueline Pan, Emily McReynolds, Miranda Bogen,
  Pascale Fung, and Cristian~Canton Ferrer.
\newblock Casual conversations v2: Designing a large consent-driven dataset to
  measure algorithmic bias and robustness, 2022.

\bibitem{DBLP:journals/corr/abs-2104-02821}
Caner Hazirbas, Joanna Bitton, Brian Dolhansky, Jacqueline Pan, Albert Gordo,
  and Cristian Canton{-}Ferrer.
\newblock Towards measuring fairness in {AI:} the casual conversations dataset.
\newblock {\em CoRR}, abs/2104.02821, 2021.

\bibitem{he2017mask}
Kaiming He, Georgia Gkioxari, Piotr Doll{\'a}r, and Ross Girshick.
\newblock Mask r-cnn.
\newblock In {\em Proceedings of the IEEE international conference on computer
  vision}, pages 2961--2969, 2017.

\bibitem{hendrycks2018benchmarking}
Dan Hendrycks and Thomas~G Dietterich.
\newblock Benchmarking neural network robustness to common corruptions and
  surface variations.
\newblock {\em arXiv preprint arXiv:1807.01697}, 2018.

\bibitem{10.2307/3088935}
Mark~E. Hill.
\newblock Race of the interviewer and perception of skin color: Evidence from
  the multi-city study of urban inequality.
\newblock {\em American Sociological Review}, 67(1):99--108, 2002.

\bibitem{holland2018dataset}
Sarah Holland, Ahmed Hosny, Sarah Newman, Joshua Joseph, and Kasia Chmielinski.
\newblock The dataset nutrition label: A framework to drive higher data quality
  standards.
\newblock {\em arXiv preprint arXiv:1805.03677}, 2018.

\bibitem{holland2020dataset}
Sarah Holland, Ahmed Hosny, Sarah Newman, Joshua Joseph, and Kasia Chmielinski.
\newblock The dataset nutrition label.
\newblock {\em Data Protection and Privacy}, 12(12):1, 2020.

\bibitem{apple}
Laura Hollink, Aysenur Bilgin, and Jacco van Ossenbruggen.
\newblock Is it a fruit, an apple or a granny smith? predicting the basic level
  in a concept hierarchy, 2019.

\bibitem{DBLP:journals/corr/HosangBDS15}
Jan~Hendrik Hosang, Rodrigo Benenson, Piotr Doll{\'{a}}r, and Bernt Schiele.
\newblock What makes for effective detection proposals?
\newblock {\em CoRR}, abs/1502.05082, 2015.

\bibitem{DBLP:journals/corr/abs-2106-11240}
John~J. Howard, Yevgeniy~B. Sirotin, Jerry~L. Tipton, and Arun~R. Vemury.
\newblock Reliability and validity of image-based and self-reported skin
  phenotype metrics.
\newblock {\em CoRR}, abs/2106.11240, 2021.

\bibitem{hutchinson2021towards}
Ben Hutchinson, Andrew Smart, Alex Hanna, Emily Denton, Christina Greer, Oddur
  Kjartansson, Parker Barnes, and Margaret Mitchell.
\newblock Towards accountability for machine learning datasets: Practices from
  software engineering and infrastructure.
\newblock In {\em Proceedings of the 2021 ACM Conference on Fairness,
  Accountability, and Transparency}, pages 560--575, 2021.

\bibitem{idrissi2022imagenet}
Badr~Youbi Idrissi, Diane Bouchacourt, Randall Balestriero, Ivan Evtimov, Caner
  Hazirbas, Nicolas Ballas, Pascal Vincent, Michal Drozdzal, David Lopez-Paz,
  and Mark Ibrahim.
\newblock Imagenet-x: Understanding model mistakes with factor of variation
  annotations.
\newblock {\em arXiv preprint arXiv:2211.01866}, 2022.

\bibitem{jo2020lessons}
Eun~Seo Jo and Timnit Gebru.
\newblock Lessons from archives: Strategies for collecting sociocultural data
  in machine learning.
\newblock In {\em Proceedings of the 2020 conference on fairness,
  accountability, and transparency}, pages 306--316, 2020.

\bibitem{kakumanu2007survey}
Praveen Kakumanu, Sokratis Makrogiannis, and Nikolaos Bourbakis.
\newblock A survey of skin-color modeling and detection methods.
\newblock {\em Pattern recognition}, 40(3):1106--1122, 2007.

\bibitem{kapania2023hunt}
Shivani Kapania, Ding Wang, and Alex Taylor.
\newblock A hunt for the snark: Annotator diversity in data practices.
\newblock 2023.

\bibitem{karkkainen2021fairface}
Kimmo Karkkainen and Jungseock Joo.
\newblock Fairface: Face attribute dataset for balanced race, gender, and age
  for bias measurement and mitigation.
\newblock In {\em Proceedings of the IEEE/CVF Winter Conference on Applications
  of Computer Vision}, pages 1548--1558, 2021.

\bibitem{khan2021one}
Zaid Khan and Yun Fu.
\newblock One label, one billion faces: Usage and consistency of racial
  categories in computer vision.
\newblock In {\em Proceedings of the 2021 acm conference on fairness,
  accountability, and transparency}, pages 587--597, 2021.

\bibitem{segment_anything}
Alexander Kirillov, Eric Mintun, Nikhila Ravi, Hanzi Mao, Chloe Rolland, Laura
  Gustafson, Tete Xiao, Spencer Whitehead, Alexander~C Berg, Wan-Yen Lo, et~al.
\newblock Segment anything.
\newblock {\em arXiv preprint arXiv:2304.02643}, 2023.

\bibitem{kleider2017black}
Heather~M Kleider-Offutt, Alesha~D Bond, and Shanna~EA Hegerty.
\newblock Black stereotypical features: When a face type can get you in
  trouble.
\newblock {\em Current Directions in Psychological Science}, 26(1):28--33,
  2017.

\bibitem{koch2reduced}
Bernard Koch, Emily Denton, Alex Hanna, and Jacob~Gates Foster.
\newblock Reduced, reused and recycled: The life of a dataset in machine
  learning research.
\newblock In {\em Thirty-fifth Conference on Neural Information Processing
  Systems Datasets and Benchmarks Track (Round 2)}.

\bibitem{kuznetsova2020open}
Alina Kuznetsova, Hassan Rom, Neil Alldrin, Jasper Uijlings, Ivan Krasin, Jordi
  Pont-Tuset, Shahab Kamali, Stefan Popov, Matteo Malloci, Alexander
  Kolesnikov, et~al.
\newblock The open images dataset v4.
\newblock {\em International Journal of Computer Vision}, 128(7):1956--1981,
  2020.

\bibitem{lin2014microsoft}
Tsung-Yi Lin, Michael Maire, Serge Belongie, James Hays, Pietro Perona, Deva
  Ramanan, Piotr Doll{\'a}r, and C~Lawrence Zitnick.
\newblock Microsoft coco: Common objects in context.
\newblock In {\em European conference on computer vision}, pages 740--755.
  Springer, 2014.

\bibitem{liu2015celeba}
Ziwei Liu, Ping Luo, Xiaogang Wang, and Xiaoou Tang.
\newblock Deep learning face attributes in the wild.
\newblock In {\em Proceedings of the IEEE international conference on computer
  vision}, pages 3730--3738, 2015.

\bibitem{loper2002nltk}
Edward Loper and Steven Bird.
\newblock Nltk: The natural language toolkit.
\newblock {\em arXiv preprint cs/0205028}, 2002.

\bibitem{meister2022gender}
Nicole Meister, Dora Zhao, Angelina Wang, Vikram~V Ramaswamy, Ruth Fong, and
  Olga Russakovsky.
\newblock Gender artifacts in visual datasets.
\newblock {\em arXiv preprint arXiv:2206.09191}, 2022.

\bibitem{miceli2021documenting}
Milagros Miceli, Tianling Yang, Laurens Naudts, Martin Schuessler, Diana
  Serbanescu, and Alex Hanna.
\newblock Documenting computer vision datasets: an invitation to reflexive data
  practices.
\newblock In {\em Proceedings of the 2021 ACM Conference on Fairness,
  Accountability, and Transparency}, pages 161--172, 2021.

\bibitem{miller1995wordnet}
George~A Miller.
\newblock Wordnet: a lexical database for english.
\newblock {\em Communications of the ACM}, 38(11):39--41, 1995.

\bibitem{mitchell2019model}
Margaret Mitchell, Simone Wu, Andrew Zaldivar, Parker Barnes, Lucy Vasserman,
  Ben Hutchinson, Elena Spitzer, Inioluwa~Deborah Raji, and Timnit Gebru.
\newblock Model cards for model reporting.
\newblock In {\em Proceedings of the conference on fairness, accountability,
  and transparency}, pages 220--229, 2019.

\bibitem{Monk_2019}
Ellis Monk.
\newblock Monk skin tone scale, 2019.

\bibitem{nosek2002harvesting}
Brian~A Nosek, Mahzarin~R Banaji, and Anthony~G Greenwald.
\newblock Harvesting implicit group attitudes and beliefs from a demonstration
  web site.
\newblock {\em Group Dynamics: Theory, Research, and Practice}, 6(1):101, 2002.

\bibitem{10.1111/bjd.19932}
U.K. Okoji, S.C. Taylor, and J.B. Lipoff.
\newblock {Equity in skin typing: why it is time to replace the Fitzpatrick
  scale}.
\newblock {\em British Journal of Dermatology}, 185(1):198--199, 07 2021.

\bibitem{paullada2021data}
Amandalynne Paullada, Inioluwa~Deborah Raji, Emily~M Bender, Emily Denton, and
  Alex Hanna.
\newblock Data and its (dis) contents: A survey of dataset development and use
  in machine learning research.
\newblock {\em Patterns}, 2(11):100336, 2021.

\bibitem{prabhu2020large}
VU Prabhu and A Birhane.
\newblock Large datasets: A pyrrhic win for computer vision.
\newblock {\em arXiv preprint arXiv:2006.16923}, 3, 2020.

\bibitem{prendergast2016case}
Heather~M Prendergast.
\newblock Case 10: West indian/caribbean.
\newblock {\em Diversity and Inclusion in Quality Patient Care}, pages
  253--258, 2016.

\bibitem{pushkarna2022data}
Mahima Pushkarna, Andrew Zaldivar, and Oddur Kjartansson.
\newblock Data cards: Purposeful and transparent dataset documentation for
  responsible ai.
\newblock In {\em 2022 ACM Conference on Fairness, Accountability, and
  Transparency}, pages 1776--1826, 2022.

\bibitem{radford2021learning}
Alec Radford, Jong~Wook Kim, Chris Hallacy, Aditya Ramesh, Gabriel Goh,
  Sandhini Agarwal, Girish Sastry, Amanda Askell, Pamela Mishkin, Jack Clark,
  et~al.
\newblock Learning transferable visual models from natural language
  supervision.
\newblock In {\em International conference on machine learning}, pages
  8748--8763. PMLR, 2021.

\bibitem{ren2015faster}
Shaoqing Ren, Kaiming He, Ross Girshick, and Jian Sun.
\newblock Faster r-cnn: Towards real-time object detection with region proposal
  networks.
\newblock {\em Advances in neural information processing systems}, 28, 2015.

\bibitem{ridnik2021imagenetk}
Tal Ridnik, Emanuel Ben-Baruch, Asaf Noy, and Lihi Zelnik-Manor.
\newblock Imagenet-21k pretraining for the masses.
\newblock In {\em Thirty-fifth Conference on Neural Information Processing
  Systems Datasets and Benchmarks Track (Round 1)}, 2021.

\bibitem{rojas2022the}
William A~Gaviria Rojas, Sudnya Diamos, Keertan~Ranjan Kini, David Kanter,
  Vijay~Janapa Reddi, and Cody Coleman.
\newblock The dollar street dataset: Images representing the geographic and
  socioeconomic diversity of the world.
\newblock In {\em Thirty-sixth Conference on Neural Information Processing
  Systems Datasets and Benchmarks Track}, 2022.

\bibitem{49953}
Nithya Sambasivan, Shivani Kapania, Hannah Highfill, Diana Akrong,
  Praveen~Kumar Paritosh, and Lora~Mois Aroyo.
\newblock "everyone wants to do the model work, not the data work": Data
  cascades in high-stakes ai.
\newblock 2021.

\bibitem{scheuerman2021datasets}
Morgan~Klaus Scheuerman, Alex Hanna, and Emily Denton.
\newblock Do datasets have politics? disciplinary values in computer vision
  dataset development.
\newblock {\em Proceedings of the ACM on Human-Computer Interaction},
  5(CSCW2):1--37, 2021.

\bibitem{10.1145/3392866}
Morgan~Klaus Scheuerman, Kandrea Wade, Caitlin Lustig, and Jed~R. Brubaker.
\newblock How we've taught algorithms to see identity: Constructing race and
  gender in image databases for facial analysis.
\newblock {\em Proc. ACM Hum.-Comput. Interact.}, 4(CSCW1), may 2020.

\bibitem{MIAP}
Candice Schumann, Susanna Ricco, Utsav Prabhu, Vittorio Ferrari, and Caroline
  Pantofaru.
\newblock A step toward more inclusive people annotations for fairness.
\newblock In {\em Proceedings of the 2021 {AAAI}/{ACM} Conference on {AI},
  Ethics, and Society}. {ACM}, jul 2021.

\bibitem{shankar2017no}
Shreya Shankar, Yoni Halpern, Eric Breck, James Atwood, Jimbo Wilson, and D
  Sculley.
\newblock No classification without representation: Assessing geodiversity
  issues in open data sets for the developing world.
\newblock {\em arXiv preprint arXiv:1711.08536}, 2017.

\bibitem{singh2022revisiting}
Mannat Singh, Laura Gustafson, Aaron Adcock, Vinicius de Freitas~Reis, Bugra
  Gedik, Raj~Prateek Kosaraju, Dhruv Mahajan, Ross Girshick, Piotr Doll{\'a}r,
  and Laurens van~der Maaten.
\newblock Revisiting weakly supervised pre-training of visual perception
  models.
\newblock In {\em Proceedings of the IEEE/CVF Conference on Computer Vision and
  Pattern Recognition}, pages 804--814, 2022.

\bibitem{steed2021image}
Ryan Steed and Aylin Caliskan.
\newblock Image representations learned with unsupervised pre-training contain
  human-like biases.
\newblock In {\em Proceedings of the 2021 ACM conference on fairness,
  accountability, and transparency}, pages 701--713, 2021.

\bibitem{DBLP:journals/corr/abs-1711-11443}
Pierre Stock and Moustapha Ciss{\'{e}}.
\newblock Convnets and imagenet beyond accuracy: Explanations, bias detection,
  adversarial examples and model criticism.
\newblock {\em CoRR}, abs/1711.11443, 2017.

\bibitem{sudhakar2023icon}
Sruthi Sudhakar, Viraj Prabhu, Olga Russakovsky, and Judy Hoffman.
\newblock Icon $^{2}$: Reliably benchmarking predictive inequity in object
  detection.
\newblock {\em arXiv preprint arXiv:2306.04482}, 2023.

\bibitem{tong2020investigating}
Schrasing Tong and Lalana Kagal.
\newblock Investigating bias in image classification using model explanations.
\newblock {\em arXiv preprint arXiv:2012.05463}, 2020.

\bibitem{wang2022revise}
Angelina Wang, Alexander Liu, Ryan Zhang, Anat Kleiman, Leslie Kim, Dora Zhao,
  Iroha Shirai, Arvind Narayanan, and Olga Russakovsky.
\newblock Revise: A tool for measuring and mitigating bias in visual datasets.
\newblock {\em International Journal of Computer Vision}, 130(7):1790--1810,
  2022.

\bibitem{DirectionalBiasAmp}
Angelina Wang and Olga Russakovsky.
\newblock Directional bias amplification.
\newblock {\em CoRR}, abs/2102.12594, 2021.

\bibitem{wang2022unifying}
Peng Wang, An Yang, Rui Men, Junyang Lin, Shuai Bai, Zhikang Li, Jianxin Ma,
  Chang Zhou, Jingren Zhou, and Hongxia Yang.
\newblock Unifying architectures, tasks, and modalities through a simple
  sequence-to-sequence learning framework.
\newblock {\em arXiv preprint arXiv:2202.03052}, 2022.

\bibitem{BDD100k_skintone}
Benjamin Wilson, Judy Hoffman, and Jamie Morgenstern.
\newblock Predictive inequity in object detection, 2019.

\bibitem{wolf2020transformers}
Thomas Wolf, Lysandre Debut, Victor Sanh, Julien Chaumond, Clement Delangue,
  Anthony Moi, Pierric Cistac, Tim Rault, R{\'e}mi Louf, Morgan Funtowicz,
  et~al.
\newblock Transformers: State-of-the-art natural language processing.
\newblock In {\em Proceedings of the 2020 conference on empirical methods in
  natural language processing: system demonstrations}, pages 38--45, 2020.

\bibitem{TFD}
Kaiyu Yang, Klint Qinami, Li Fei-Fei, Jia Deng, and Olga Russakovsky.
\newblock Towards fairer datasets.
\newblock In {\em Proceedings of the 2020 Conference on Fairness,
  Accountability, and Transparency}. {ACM}, jan 2020.

\bibitem{bdd100k}
Fisher Yu, Haofeng Chen, Xin Wang, Wenqi Xian, Yingying Chen, Fangchen Liu,
  Vashisht Madhavan, and Trevor Darrell.
\newblock Bdd100k: A diverse driving dataset for heterogeneous multitask
  learning.
\newblock In {\em IEEE/CVF Conference on Computer Vision and Pattern
  Recognition (CVPR)}, June 2020.

\bibitem{zhifei2017cvpr}
Song~Yang Zhang, Zhifei and Hairong Qi.
\newblock Age progression/regression by conditional adversarial autoencoder.
\newblock In {\em IEEE Conference on Computer Vision and Pattern Recognition
  (CVPR)}. IEEE, 2017.

\bibitem{https://doi.org/10.48550/arxiv.2210.11924}
Dora Zhao, Jerone T.~A. Andrews, and Alice Xiang.
\newblock Men also do laundry: Multi-attribute bias amplification, 2022.

\bibitem{zhao2021understanding}
Dora Zhao, Angelina Wang, and Olga Russakovsky.
\newblock Understanding and evaluating racial biases in image captioning.
\newblock In {\em Proceedings of the IEEE/CVF International Conference on
  Computer Vision}, pages 14830--14840, 2021.

\bibitem{zhao2017men}
Jieyu Zhao, Tianlu Wang, Mark Yatskar, Vicente Ordonez, and Kai-Wei Chang.
\newblock Men also like shopping: Reducing gender bias amplification using
  corpus-level constraints.
\newblock {\em arXiv preprint arXiv:1707.09457}, 2017.

\bibitem{zhou2022detecting}
Xingyi Zhou, Rohit Girdhar, Armand Joulin, Philipp Kr{\"a}henb{\"u}hl, and
  Ishan Misra.
\newblock Detecting twenty-thousand classes using image-level supervision.
\newblock In {\em Computer Vision--ECCV 2022: 17th European Conference, Tel
  Aviv, Israel, October 23--27, 2022, Proceedings, Part IX}, pages 350--368.
  Springer, 2022.

\end{thebibliography}
}

\
\clearpage
\newpage
\begin{appendix}
\section{Appendix}
\label{appendix}
\subsection{Category Selection}
\begin{figure*}[h]
\centering
\includegraphics[width=0.8\linewidth,angle=-86.5]{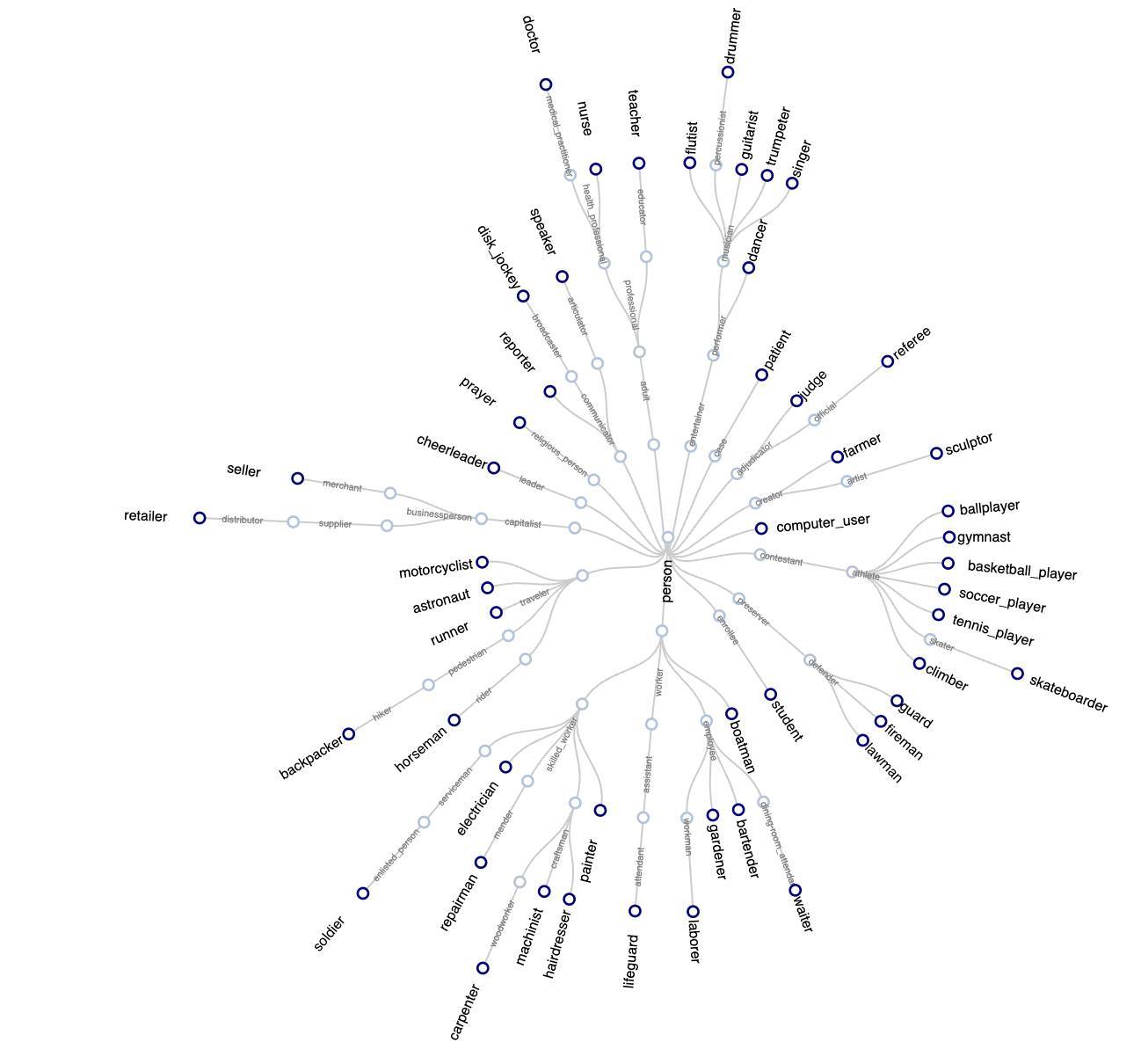}
\caption{WordNet hierarchy of the FACET classes in relation to the \texttt{Person} synset. Classes are mapped to the \texttt{Person} synset (center) by their hypernyms (parents). Classes (leaves) are marked in blue. Grey nodes correspond to an intermediate hyponyms.}
\label{fig:wordnet_facet}
\end{figure*}

\paragraph{Imageability and Salience of the Categories}
When choosing the categories for \datasetabbr, we considered the ``imageability'' of our concepts, from \cite{TFD}.
However, we found that this did not transfer well to our use case.
First, we found that many of the `highly imageable' concepts include classes directly related to a demographic attribute. (For instance, \textit{black woman ($n09637339$)} has an imageability score of 5 out of 5.)
Additionally, many highly imageable concepts are abstract, meaning they are easy to imagine but hard to classify. As an example, it is easy to imagine what the concept \texttt{mother} may look like, but it is hard to determine if someone is a ``mother" from a photo.
(\textit{Is any person perceived with having stereotypically feminine attributes with a child in a photo presumed to be a mother?})
\paragraph{Class Hierarchy and Representation}
We show the full connection of our chosen concepts in WordNet in their relation to the \texttt{Person} synset. 
Figure \ref{fig:wordnet_facet} shows the full connection of our chosen concepts in WordNet in their relation to the \texttt{Person} synset. All relevant sub-trees and intermediate synsets are shown. We can see that many of the classes in \datasetabbr share the same parent node. We also note that no class in \datasetabbr is a direct descendant of another class. This demonstrates that there is no overlap between classes. Table \ref{table:class_all} shows the representation of each class in the evaluation set.

\subsection{Annotation Pipeline}
\label{app:benchmark_decisions}
We describe in more detail the annotation pipeline we use for FACET.

\begin{figure}[h]
    \includegraphics[width=0.8\linewidth]{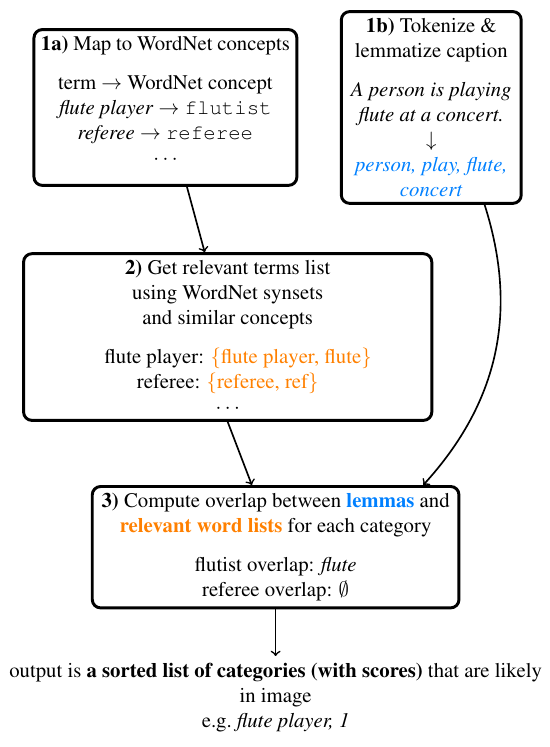}

   \caption{\textbf{Label annotation pipeline:} The preprocessing steps before beginning the annotation pipeline. \textbf{In 1a)} we map all of the person-related classes to concepts in WordNet. We denote WordNet concepts in a different font. (See Section \ref{bench-method} for a full description on WordNet concepts and synsets). \textbf{In 1b)} we tokenize and lemmatize the captions to produce a list of \textbf{\lightBlue lemmas}. \textbf{In 2)} for each of the \numpersonclasses categories, we generate a \textbf{\orange relevant words list} using WordNet synsets. Lastly \textbf{in 3)}  we compute overlap between the \textbf{\lightBlue lemmas} and \textbf{\orange relevant words list} and select images to annotate which have high overlap.}
\label{fig:pipeline}
\end{figure}

\begin{figure}[h]
    \centering
    \includegraphics[width=0.8\linewidth]{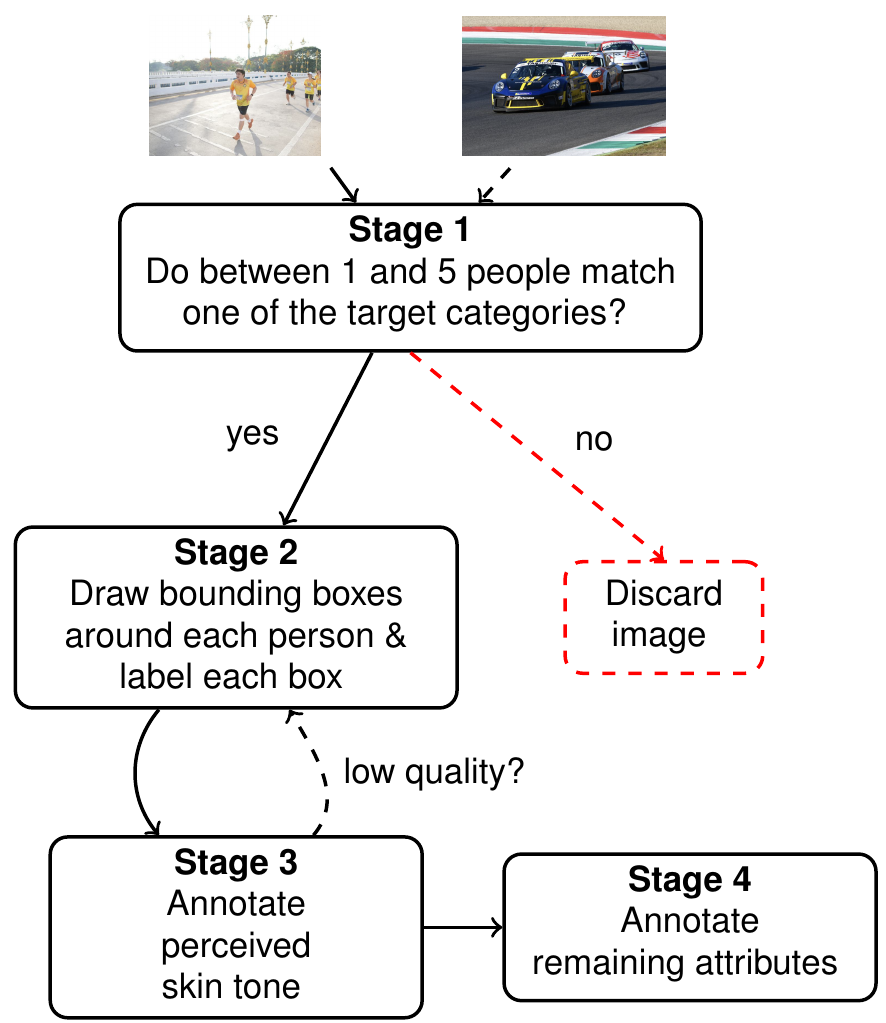}

    \caption{
    \textbf{Image annotation pipeline:} The four stages of the main annotation pipeline. The image on the left can be fully annotated; the image on the right does not contain the target categories and gets excluded after Stage 1. Dashed lines show paths that do not advance to the next stage.
    }
    \label{fig:ann_pipeline}
\end{figure}

\subsubsection{Annotation Pipeline Design}
\label{app:sec:pipeline}
\paragraph{Preprocessing} Figure \ref{fig:pipeline} shows the pre-processing steps of the captions to create the candidate set of images to annotate. We `score' each caption for each category based on the overlap of relevant words for the category and caption. We sample captions with the highest `score' per category. We choose the candidate images
for \datasetabbr from a set of roughly 6 million images.

We select a starting set of images for annotation such that we expect the portion of images that pass stage 1 to be roughly class balanced. To approximate the probability that images with overlap per category are true positives, we sample 50 images per category and annotate the true positives. We use this frequency to determine how much to over sample a specific category. As we continue the annotation process, for additional rounds, we sample images with overlap based on the categories that are under-represented in the dataset thus far. We note that many categories did not have enough images with matching relevant words and as such we did not achieve equal representation of all categories.

\paragraph{Annotation Stages}
Figure \ref{fig:ann_pipeline} shows the four separate annotation tasks of the main annotation pipeline. Breaking the annotation process into multiple sub-tasks allows for more fine-grained control. For Stage 1, we focus on speed, and ask annotators to spend little time per task.  To increase speed, we group multiple images with the same target categories into the a single task with a default value of \textit{0 people match the categories}, and ask the annotator to label each image. We separate Stages 3 and 4, so that we can gather multiple annotations for apparent skin tone only. We separate these stages from Stage 2 to simplify the task for annotators, such that they only need consider the perceived demographic attributes for one person at a time. Additionally, this allows the annotators in later stages to quality assure the annotations from earlier stages, as described in Section \ref{app:ann_qa}.

\paragraph{Mask Annotations}
We collect labels for \datasourcenocite\ mask annotations separately after completing the annotation pipeline. 
First, we select candidate people from FACET with attempts made to balance the number of people per demographic group. Next, we select a candidate set of masks to annotate by collecting the set of masks inside the bounding boxes for these people. For each mask and FACET bounding box in which it resides, we asked annotators if the mask corresponds to \textit{the person's body, the person's hair} or \textit{an item of clothing on the person} for a given person, as denoted by a bounding box. Annotators did not make any modifications to the masks, e.g. change the shape. Annotators were told to only select a class if the mask covers the entire item; masks for a portion of the person, or part of item of clothing were not labeled. Additionally, annotators were only told to select a class if it met the label for the person described by the bounding box; masks for people, hair, clothing inside of the bounding box but belonging to a different person were not labeled. Thus, each mask is attached to a specific person in FACET. The breakdown of the masks per image is is given in \ref{app:sec:stat}. The breakdown of masks per given demographic or additional attribute is given in Table \ref{app:mask_stats}.

\subsection{Annotation Quality Assurance (QA)}

\label{app:ann_procedure}
\subsubsection{Annotator Quality Assurance using Training}
Before completing any annotations used in FACET, annotators 
were trained for each stage separately. We trained annotators by giving them a sample set of tasks and comparing their annotations to a known golden set.
For Stages 1, 3 and 4 (image filtering, \skintone annotation, other perceived attributes annotation), annotators passed the training step if the recall of their annotations compared to a fixed golden set was above a quality threshold.
This threshold was set for each stage depending on the difficulty of the task. For Stage 2 (drawing bounding boxes), QA was done per annotator to assess the quality of boxes. We provided feedback to annotators individually and only graduated the annotators once they addressed the feedback. 
A manual IoU threshold of 0.85 between an annotator and the golden set was used. Annotators under that threshold were not manually reviewed, as we found that this correlated with extremely poor box quality, and these annotators did not graduate training. Before feedback, we noticed that many annotators were drawing bounding boxes that included objects the person was holding (\textit{e.g. guitar}) as opposed to tightly around the person. After manual review and feedback, the quality of the annotations was much higher and consistent.
\subsubsection{Annotation Quality Assurance using Multi-Review and Quality Checks}
\label{app:ann_qa}
In addition to implementing a multi-review process for the \skintone annotations of each target person as discussed in Section \ref{stage2}, we used Stage 3 to QA the bounding boxes drawn by the annotators. The annotators in Stage 3 were asked whether the bounding box for the person in the task was drawn tightly around the person.
If -- for any bounding box in the image -- any of the three annotators marked that the bounding box was not tight, the image was placed back in Stage 2 of the pipeline to be re-annotated.

\subsection{Dataset Statistics and Breakdown}
\label{app:data_stats}

\subsubsection{Attribute Representation}
We detail the attribute breakdown for the remaining annotations in FACET. Table \ref{app:table:demographic_perc} details the statistics for the remaining person annotations. Table \ref{app:table:robustness} shows the results of the robustness annotations with breakdowns on occlusion level and lighting condition.

\newcommand{\STAB}[1]{\begin{tabular}{@{}c@{}}#1\end{tabular}}

\begin{table}[!h]
  \centering
  \begin{tabular}{c l|r r r r}
     & \multicolumn{1}{c|}{} & \multicolumn{1}{c}{\it people} & \multicolumn{1}{c}{\%} & \multicolumn{1}{c}{\it images} & \multicolumn{1}{c|}{\%}  \\ \hline
     \multirow{7}{*}{\STAB{\rotatebox[origin=c]{90}{\parbox{1.5cm}{\it \it Hair color}}}} 
& black &17k & 34\%  &  13k &   42\% \\
& blonde &  3k &  6\%  &  3k & 8\%\\
& brown &  11k &  22\% & 9k & 29\% \\
& red/orange & 547  &  1\%  &  518  &  2\% \\
& colored &269  & 1\%  &265  &  1\%  \\
& grey & 2k &  4\% &   2k &  6\% \\
& unknown &  20k & 40\% &  15k & 46\%  \\
\hline
\multirow{7}{*}{\STAB{\rotatebox[origin=c]{90}{\parbox{2cm}{\it Hair type}}}} 
& wavy & 9k & 19\% &  8k & 26\%\\
& curly & 761  &   2\%&  735  &  2\%  \\
& straight &  19k & 37\% & 15k &  47\% \\
& coily & 458  & 1\%  & 435  &  1\% \\
& dreadlocks & 296  & 1\% &  282  & 1\%  \\
& bald &  1k &  2\%  & 965  &  3\%\\
& unknown & 23k & 45\%  & 16k &  52\% \\
\hline
\multirow{6}{*}{\STAB{\rotatebox[origin=c]{90}{\parbox{2cm}{\it Additional Annotations}}}} 
& eyeware &  5k & 11\% &   5k & 15\% \\
& headscarf &  2k &  5\%  &  2k &  6\%  \\
& tattoo & 705  & 1\%  & 672  & 2\%  \\
& cap &   14k &  29\% &  10k &  33\%  \\
& facial-hair &  6k & 12\% & 5k &  17\%\\
& mask &  3k & 6\% & 2k & 7\% \\
\hline
\end{tabular}
\caption{Statistics on the remaining person attributes: \textit{hair color, hair type, presence of additional features} in \datasetabbrnospace. Annotators could mark multiple hair colors and types for a single person.}
\vspace{2em}
\label{app:table:demographic_perc}
\end{table}

\begin{table}[!h]
  \centering
  \begin{tabular}{c l|r r r r}
    
     & \multicolumn{1}{c|}{\it label} & \multicolumn{1}{c}{\it people} & \multicolumn{1}{c}{\%} & \multicolumn{1}{c}{\it images} & \multicolumn{1}{c|}{\%}  \\ \hline
     \multirow{5}{*}{\STAB{\rotatebox[origin=c]{90}{\parbox{1.5cm}{\centering \textit{Lighting Condition}}}}} 
& overexposed & 941  &  2\% &   890  &         3\% \\
& well-lit & 40k &  80\% & 27k &  85\%  \\
& dimly-lit &  11k & 22\%&  9k &  28\%\\
& underexposed &  1k &  3\%  & 1k & 4\% \\
& unknown &   878  & 2\% & 849  &   3\%\\
\hline
\multirow{4}{*}{\STAB{\rotatebox[origin=c]{90}{\parbox{1cm}{\centering \textit{Visibility}}}}}
& minimal &  7k &        15\% & 7k &        21\% \\
& face &   15k &        30\% & 12k &        38\% \\
& torso &    36k &        73\% &  25k &        78\% \\

\end{tabular}
\caption{Robustness annotations.}
\label{app:table:robustness}
\end{table}

\subsubsection{Image statistics}
\label{app:sec:stat}
\begin{figure}
  \begin{centering}
 \resizebox{0.48\columnwidth}{1.3in}{
  \begin{tikzpicture}
    \begin{axis}[ybar interval, compat=newest, ymin=0,ymax=75, xmin=1,xmax=8,ylabel=\LARGE percent of images,
    xlabel= \LARGE number of people,]
    \addplot coordinates {(1, 68) (2, 19) (3, 7) (4, 3) (5, 2) (6, 1) (7, 0)};
    \end{axis}
  \end{tikzpicture}
  }  \caption{\label{app:fig:ppl_hist}Histogram of number of people per image in \datasetabbrnospace.}
  \vspace{3em}
  \end{centering}
  \end{figure}
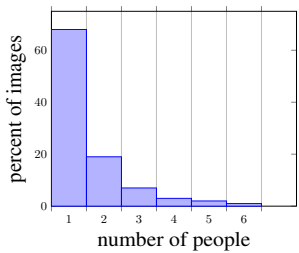

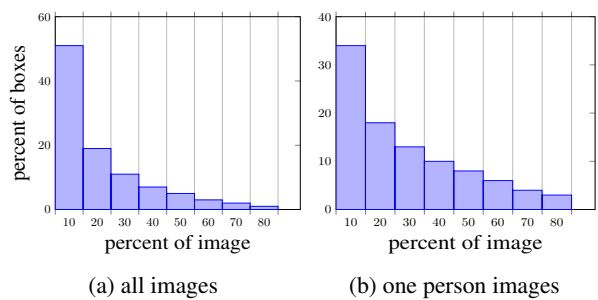
\begin{figure}
\centering
    \begin{subfigure}[b]{0.48\columnwidth}
        \centering
        \resizebox{\linewidth}{1.3in}{
  \begin{tikzpicture}
    \begin{axis}[ybar interval, compat=newest, ymin=0,ymax=60, xmin=10,xtick=data,
    xlabel= \LARGE percent of image,  ylabel= \LARGE percent of boxes,xtick=data,]]
    \addplot coordinates {(10.0, 51) (20.0, 19) (30.0, 11) (40.0, 7) (50.0, 5) (60.0, 3) (70.0, 2) (80.0, 1) (90.0, 1)};
    \end{axis}
  \end{tikzpicture}
        }
        \caption{all images}
        \label{fig:subfig8}
    \end{subfigure}
    \begin{subfigure}[b]{0.46\columnwidth}
    \centering
        \resizebox{\linewidth}{1.3in}{
  \begin{tikzpicture}
    \begin{axis}[ybar interval, compat=newest, ymin=0,ymax=40, xmin=10, xtick=data,
    xlabel= \LARGE percent of image, ]
    \addplot coordinates {(10.0, 34) (20.0, 18) (30.0, 13) (40.0, 10) (50.0, 8) (60.0, 6) (70.0, 4) (80.0, 3) (90.0, 2)};
    \end{axis}
  \end{tikzpicture}
  }
        \caption{one person images}   
    \end{subfigure}
    \vspace{1em}
    \caption{Histogram of person bounding box size as a percentage of total image size.}
    \vspace{3em}
    \label{app:fig:person_boxes}
\end{figure}

\vspace{-0.15em}
We measure the statistics of images beyond specific attributes.
Figure \ref{app:fig:ppl_hist} shows the number of annotated people per image; less than one third of the images contain more than one person. Figure \ref{app:fig:person_boxes} shows the person box size as a fraction of total image size, broken down by the number of people in the image. All images in FACET are used for detection. Images with only one person are used for classification and visual grounding. \textbf{For masks, the \masksizemini labeled masks span 18k people in 17k images of FACET. Each person with associated labeled masks has an average of 4 masks.}

\subsection{Evaluation}
\label{app:evaluation}

\subsubsection{Dataset Setup}
\begin{itemize}
    \item For image classification, we limit the evaluation to examples in FACET that only contain one person. This helps alleviate ambiguities in performance. With this setup, we can consider the performance of the model on an image equivalent to performance of the model on the image for a specific set of attributes. There are 21k images in \datasetabbr that meet this criteria.
    \item For person and open world detection, we use all examples in FACET.
    \item For person segmentation, and the corresponding person detection baseline, we only use images and people inside each image that had a \texttt{person} mask - 11k people.
    \item For visual grounding, we only use examples in FACET with one person, as OFA predicts only one bounding box.
\end{itemize}

\begin{figure}[h]
    \centering
    \includegraphics[width=0.97\columnwidth]{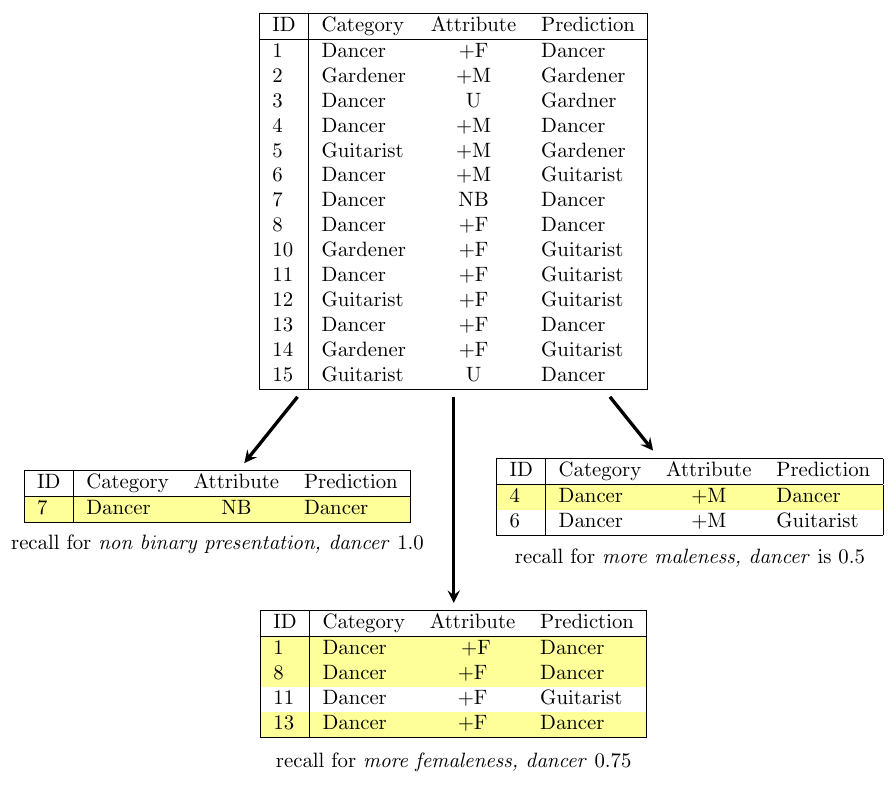}
    \caption{Example of how we score classification models for \datasetabbrnospace.
    }
    \label{fig:recall}
\end{figure}

\subsubsection{Choice of Metric}

We choose to focus on recall as it allows us to only consider examples with a specific demographic attribute or set of attributes. We choose to avoid a metric that would take into account false positives, as for some evaluations it is not clear what a false positive would mean. For example, for person detection, it is not obvious which demographic attribute a false positive would correspond to. \textit{What demographic attributes would we consider a predicted false positive person to have?} While it might make sense for images with only one person to assume they had the same demographic attributes as the ground truth person in the photo, it is even less clear what the correct assumption would be to make if there were multiple people in the photo. To avoid this ambiguity, we focus on recall.

\paragraph{Classification}
We compare performance on a per-class basis, as overall performance metrics can hide disparities -- \ie the model could have large biases but in opposite directions for two classes, which would yield a overall performance disparity of 0.
We also don't want the metric to be influenced by the prevalence of the class for the group. We focus on the recall (R@1) for the group and class for our evaluation. This is equivalent to the accuracy for the specific \texttt{(class, attribute)} pair. We do not account for true negatives or false positives. Figure \ref{fig:recall} visualizes our metric. We note that there are multiple approaches to calculating a metric per class -- \eg we could also look at the accuracy for the class when looking at all examples of the protected group, which is why detail the specifics of our considered metric.

\paragraph{Alignment with traditional fairness metrics}
The difference in recall we measure is equivalent to \textit{equality of opportunity} \cite{hardt2016equality} -- larger differences in recall are further from equality of opportunity. \textit{Equalized odds} \cite{hardt2016equality} is an extension of this, which analyzes the true negative rate (TNR). For example for CLIP, the largest difference in TNR between people perceived as having more stereotypically male and more stereotypically female attributes is for \texttt{retailers}, with 3.8 difference in TNR. This suggests that people perceived as having more stereotypically female attributes are over-predicted as \texttt{retailers}. The next largest difference between these two groups for CLIP is for \texttt{tennis player}, with a difference in TNR of $-3.0$, suggesting that people perceived as having more stereotypically male attributes are over-predicted as \texttt{tennis players}.

\subsubsection{Classification}\label{sec:exp_setup_cls}
\paragraph{Experimental Setup}
In order to have maximum control over the experiment, we
evaluate classification models on images in FACET that only contain one annotated person.
By filtering out images with $>$ 1 person, we are left with 21k images. We look at the per-class disparities between two
groups only if both groups have at least 50 examples. We
analyze CLIP based on recall. 
\paragraph{ImageNet-21k Pretraining}
As \datasetabbr categories overlap with ImageNet classes, we can evaluate models trained on ImageNet-21k out of the box. We take the max score over the FACET classes from the ImageNet class predictions.
Table \ref{fig:class_other_attr} shows a comparison of performance discrepancies across \age for CLIP ViT B/32 and a ViT B/16 pre-trained on IN21k from \cite{ridnik2021imagenetk}.

\renewcommand{\tabcolsep}{3.5pt}
\begin{table}[h]
\centering\small
\begin{tabular}{ l l|l c c c ||l  c c c}
 & \it \multirow{2}{*}{\parbox{0.8em}{\centering\small Person\\ Class}}
& \multicolumn{4}{c||}{CLIP ViT B/32}
& \multicolumn{4}{c}{ViT B/16 IN21k}\\

& 
& \multicolumn{1}{c|}{\footnotesize \#}
& \multicolumn{1}{c|}{\footnotesize Y}
& \multicolumn{1}{c|}{\footnotesize M} 
& \multicolumn{1}{c||}{\footnotesize O}
& \multicolumn{1}{c|}{\#}
& \multicolumn{1}{c|}{\footnotesize Y}
& \multicolumn{1}{c|}{\footnotesize M}
& \multicolumn{1}{c}{\footnotesize O} \\
\hline
     \multirow{5}{*}{\STAB{\rotatebox[origin=c]{90}{\parbox{1.5cm}{\centering \footnotesize\textit{Top for CLIP}}}}} &
seller & 1 & \bf 57.5 & 72.8 & \bf 86.2  & 9 & \bf 47.2 & 53.4 & \bf 59.3  \\
& ballplayer & 2 & \bf 60.6 & \bf 75.5 & - &2 & \bf 57.6 & \bf 77.4 &  - \\
& guitarist & 3 & 70.3 & \bf 80.2 & \bf 65.5  & 10 & 45.5 & \bf 47.9 & \bf 36.4  \\
& speaker & 4 &  \bf 17.6 & 28.5 & \bf 30.6  & 4 & \bf 13.7 & 25.7 & \bf 30.6  \\
 5 & laborer & 5 & \bf 49.0 & 52.7 & \bf 61.7 & 3 & \bf 48.1 & 52.9 & \bf 66.0  \\
\hline
     \multirow{5}{*}{\STAB{\rotatebox[origin=c]{90}{\parbox{1.5cm}{\centering\footnotesize \textit{Top for ViT}}}}} &
painter & 21 & \bf 56.5 & \bf 51.0 & 53.9 & 1 & \bf 37.0 & 43.1 & \bf 57.8   \\
& ballplayer & 2 & \bf 60.6 & \bf 75.5 &  - & 2 & \bf 57.6 & \bf 77.4 &  - \\
& laborer &5 & \bf 49.0 & 52.7 & \bf 61.7  &  3 & \bf 48.1 & 52.9 & \bf 66.0  \\
& speaker & 4 & 17.6 & 28.5 & \textbf{30.6} & 4 & 13.7 & 25.7 & \textbf{30.6} \\
& guard & 7 & \textbf{44.6} & 32.9 & - & 5 & \textbf{48.5} & 31.7 &  - \\
\hline
\end{tabular}
\caption{Per-class performance for CLIP and a ViT pre-trained on ImageNet-21k. A subset of \datasetabbr classes are shown. The perceived age groups with the highest performance discrepancy
per class are bolded. (Y is \textit{young}, M is \textit{middle}, O is \textit{older}). The top five classes with the biggest discrepancies per model are shown. \# corresponds to the rank for class in terms of magnitude of the discrepancy. Lower number indicates larger discrepancy. We note that most classes are in both of the models' top 10 classes with the largest discrepancies,; 2 classes are in both models' top 5.
Recall for class and \age pairings with less than 50 samples are not reported. \\
}
\label{fig:class_other_attr}
\end{table}

\paragraph{Architecture Choice}

\renewcommand{\tabcolsep}{3.5pt}
\begin{table}[h]
\centering\small
\begin{tabular}{ l l|l c c c ||l  c c c}
 & \it \multirow{2}{*}{\parbox{0.8em}{\centering\small Person\\ Class}}
& \multicolumn{4}{c||}{ResNet IN21k}
& \multicolumn{4}{c}{ViT IN21k}\\

& 
& \multicolumn{1}{c|}{\footnotesize \#}
& \multicolumn{1}{c|}{\footnotesize Y}
& \multicolumn{1}{c|}{\footnotesize M} 
& \multicolumn{1}{c||}{\footnotesize O}
& \multicolumn{1}{c|}{\#}
& \multicolumn{1}{c|}{\footnotesize Y}
& \multicolumn{1}{c|}{\footnotesize M}
& \multicolumn{1}{c}{\footnotesize O} \\
\hline
     \multirow{5}{*}{\STAB{\rotatebox[origin=c]{90}{\parbox{1.5cm}{\centering \footnotesize\textit{Top for ResNet}}}}} & 
laborer & 1 & 35.6 & 38.1 & 55.3  & 3 & 48.1 & 52.9 & 66.0  \\
& guard & 2 & 49.5 & 30.5 &  & 6 & 48.5 & 31.7 &   \\
& painter & 3 & 38.9 & 35.9 & 53.9  &1 & 37.0 & 43.1 & 57.8  \\
& ballplayer &4 & 62.1 & 79.3 &  & 2 & 57.6 & 77.4 &   \\
& craftsman & 5 & 67.2 & 78.4 & 81.8  & 12 & 74.6 & 78.7 & 81.8  \\
\hline
     \multirow{5}{*}{\STAB{\rotatebox[origin=c]{90}{\parbox{1.5cm}{\centering\footnotesize \textit{Top for ViT}}}}} &
painter & 3 & 38.9 & 35.9 & 53.9 & 1 & \bf 37.0 & 43.1 & \bf 57.8   \\
& ballplayer & 4 & 62.1 & 79.3 &   & 2 & \bf 57.6 & \bf 77.4 &  - \\
& laborer &1 & 35.6 & 38.1 & 55.3  &  3 & \bf 48.1 & 52.9 & \bf 66.0  \\
& speaker & 15 & 20.6 & 25.9 & 24.6  & 4 & 13.7 & 25.7 & 30.6 \\
& guard & 2 & 49.5 & 30.5 &   & 5 & 48.5 & 31.7 &  - \\
\hline
\end{tabular}
\caption{Per-class performance for a ViT B/16 and ResNet-50 pre-trained on ImageNet-21k. A subset of \datasetabbr classes are shown. The perceived age groups with the highest performance discrepancy
per class are bolded. (Y is \textit{young}, M is \textit{middle}, O is \textit{older}). The top five classes with the biggest discrepancies per model are shown. \# corresponds to the rank for class in terms of magnitude of the discrepancy. Lower number indicates larger discrepancy. 
Recall for class and \age pairings with less than 50 samples are not reported. \\
}
\label{tab:vit_resnet_architecture}
\end{table}

Table \ref{tab:vit_resnet_architecture} compares performance discrepencies across architectures for models pre-trained on ImageNet-21k\cite{ridnik2021imagenetk}.

\subsubsection{Person Detection}\label{sec:exp_setup_person}
We use a pre-trained Faster R-CNN with a ResNet50 FPN backbone pretrained on COCO for person detection.

\begin{table}
  \begin{center}
  \begin{tabular}{l|c c c}
  \it Demographic Group & mAR & \multicolumn{1}{|c|}{AR$_{0.5}$} & \multicolumn{1}{c|}{AR$_{0.75}$} \\\hline
  \gender & & & \\
  \scalebox{0.9}{\hspace*{1em}-- more stereotypically maleness} &  \bf 74.4 & 97.8 & \bf 83.1 \\
  \scalebox{0.9}{\hspace*{1em}-- more stereotypically femaleness} & 72.2 & \bf 97.9 & 80.7 \\
  \scalebox{0.9}{\hspace*{1em}-- outside of gender binary} &  71.2 &\bf 97.9 & 76.8  \\
  \hline
  \hline
  \age & & & \\
  \hspace{1em}-- younger & 73.9 & 98.3 & 82.6 \\
  \hspace{1em}-- middle & 74.3 & 98.0 & 83.1 \\
  \hspace{1em}-- older & \bf 74.8 & \bf 98.5 & \bf 84.5 \\
  \hline
  
  \end{tabular}
  \caption{Average recall (AR) on FACET for a ResNet50 Faster R-CNN. Mean AR (mAR) averages across IoUs from 0.5 to 0.95 in increments of 0.05; AR$_{0.5}$ and AR$_{0.75}$ refer to IoU at 0.5 and 0.75.
  } 
  \label{fig:maskrcnn}
  \end{center}
  \label{table:st_maskrcnn}
  \end{table}

  \begin{table}
  \begin{center}
  \scalebox{1.1}{
  
      \begin{tabular}{l|c c c}
      \it Monk Skin Tone (MST) & mAR & \multicolumn{1}{|c|}{AR$_{0.5}$} & \multicolumn{1}{c|}{AR$_{0.75}$} \\\hline
      
       1 \mstone & \bf 	85.4  & 	\bf 99.0  & 	\bf 93.3 \\
       2 \msttwo & 	84.6  & 	98.8  & 	92.1 \\
       3 \mstthree & 84.4  & 	98.7  & 	91.6 \\
       4 \mstfour & 84.2  & 	98.6  & 	91.3 \\
       5 \mstfive & 84.0  & 	98.6  & 	91.2 \\
       6 \mstsix & 84.0  & 	98.7  & 	91.2 \\
       7 \mstseven & 83.8  & 	98.6  & 	91.1 \\
       8 \msteight & 84.1  & 	98.6  & 	91.5 \\
       9 \mstnine & 83.6  & 	98.6  & 	90.9 \\
       10 \mstten & 	82.8  & 	98.2  & 	90.1 \\
      \hline
      \end{tabular}
  }
  \caption{Average recall (AR) on FACET for a ResNet50-backbone DETR model. Mean AR (mAR) averages across IoUs from 0.5 to 0.95 in increments of 0.05; AR$_{0.5}$ and AR$_{0.75}$ refer to IoU at 0.5 and 0.75.
  } 
  \label{table:st_detr}
  \end{center}
  \end{table}

\paragraph{Additional Results}
Table \ref{fig:maskrcnn} shows person detection results across \gender and \agenospace. 
Table \ref{table:st_detr} shows person detection results for a DETR \cite{carion2020end} model with a ResNet50 backbone for \skintonenospace.

\subsubsection{Person Segmentation}\label{sec:exp_setup_person_deg}
We use a Mask R-CNN \cite{he2017mask} with a ResNet50 FPN backbone pretrained on COCO for person detection and instance segmentation. For this experiment, we only evaluate people in images if they have a mask annotated as \texttt{person} as well. This leaves us with 11k examples (people). For boxes, we compute the IoU of the predicted box to the human-labeled bounding box in FACET. For masks, we compute the IoU of the predicted mask to the Segment Anything-generated, annotator verified, mask in \datasourcelong. Annotators verified and labelled the mask as \texttt{person}, and were instructed only to do so if the mask was around the entire person (similar to bounding boxes in FACET). Annotators did not make any updates to the mask boundary.

\subsubsection{Open World Detection}\label{sec:exp_setup_det}
\paragraph{Experimental Setup}
We use Detic \cite{zhou2022detecting} trained on IN21-k with a SWIN-B backbone for open world detection. For the CLIP embeddings, we use the prompt `\texttt{a person who is a \{\}}' opposed to the `\texttt{a \{\}}' used in the original paper.
As we focus on recall, we do not use a confidence threshold for DETIC's predictions. Similarly we allow multiple class predictions per box. We take the top 100 predictions per image to compute AR.

\paragraph{Additional Results}
Table \ref{tab:age_both} shows the per class disparities for all classes for \agenospace.

\renewcommand{\tabcolsep}{4pt}
\begin{table*}[h]
\centering
\begin{subtable}[t]{0.44\textwidth}
\scriptsize
    \begin{tabular}{r|ccc | ccc | ccc}
        & \multicolumn{3}{c}{\small mAR}
        & \multicolumn{3}{c}{\small AR$_{0.5}$}
        & \multicolumn{3}{c}{\small AR$_{0.75}$}\\
        & \rotatebox{90}{\footnotesize young} & \rotatebox{90}{\footnotesize middle} & \rotatebox{90}{\footnotesize older}
        & \rotatebox{90}{\footnotesize young} & \rotatebox{90}{\footnotesize middle} & \rotatebox{90}{\footnotesize older}
        & \rotatebox{90}{\footnotesize young} & \rotatebox{90}{\footnotesize middle} & \rotatebox{90}{\footnotesize older}
        \\\hline
astronaut &64.0 & \bf 70.2 & - & 80.0 & \bf 85.3 & - & 60.0 & \bf 73.7 & -  \\ 
backpacker & \bf 45.4 & 42.1 & 29.8 & \textbf{55.3} & 51.7 & 35.3 & \textbf{47.4} & 44.6 & 33.3  \\ 
ballplayer & 43.8 & 45.4 & \bf 61.8 & 46.7 & 48.3 & \textbf{63.6} & 45.8 & 46.9 & \textbf{63.6}  \\ 
bartender &\textbf{81.7} & 75.4 & - & 83.3 & \textbf{85.4} & - &\textbf{ 83.3} & 80.5 & - \\ 
basketball player &61.2 & \textbf{66.7}  & - & 69.0 & \textbf{74.7 }& - & 64.9 & \textbf{ 70.4 }& -  \\ 
boatman &\textbf{69.4} & 59.9 & 64.0 & \textbf{87.7} & 79.1 & 82.3 & \textbf{77.8} & 65.3 & 68.8  \\ 
carpenter &67.9 & 64.8 & \textbf{81.7} & 71.4 & 73.5 & \textbf{91.7 }& 71.4 & 68.9 & \textbf{87.5}  \\ 
cheerleader &\textbf{13.3} & 12.7 & - & 15.6 & \textbf{13.5} & -& \textbf{14.8 }& 12.6 &-  \\ 
climber &\textbf{76.6 }& 74.4 & 67.5 & \textbf{91.9 }& 92.4 & 75.0 & \textbf{81.8} & 81.0 & 75.0  \\ 
computer user &\textbf{72.9} & 66.3 & 68.4 & \textbf{81.0} & 77.8 & 73.7 & \textbf{76.3} & 67.5 & 68.4  \\ 
craftsman &44.5 & 47.0 & \textbf{56.9} & 48.6 & 52.1 & \textbf{61.9} & 44.8 & 48.3 & \textbf{58.6 } \\ 
dancer &\textbf{77.2} & 71.1 & 75.6 & \textbf{91.4} & 85.4 & 87.5 & \textbf{83.7} & 77.5 & 78.1  \\ 
disk jockey &\textbf{77.2} & 68.4 & - &\textbf{ 82.1} & 78.8 & -& \textbf{79.1 }& 72.5 & - \\ 
doctor &74.6 & \textbf{77.4 }& 75.7 & 86.2 & \textbf{88.7} & 81.0 & 76.6 & \textbf{79.8 }& 78.6  \\ 
drummer &19.9 & 26.3 & \textbf{34.2} & 24.9 & 34.7 & \textbf{41.8 }& 19.7 & 27.6 & \textbf{35.8}  \\ 
electrician &\textbf{56.3} & 51.4 & 48.6 & \textbf{62.8 }& 62.5 & 57.1 & \textbf{62.8} & 54.4 & 57.1  \\ 
farmer &81.5 & 81.1 & \textbf{85.4} & 95.9 & 96.6 & \textbf{99.1} & 86.9 & 88.4 & \textbf{93.0 } \\ 
fireman &\textbf{86.3} & 76.4 & 76.4 & \textbf{96.2} & 90.1 & 85.7 &\textbf{ 90.4} & 82.6 & 85.7  \\ 
flutist &32.1 & 40.5 & \textbf{51.0 }& 35.4 & 47.5 & \textbf{54.8} & 35.4 & 43.7 &\textbf{ 54.8 } \\ 
gardener &82.3 & 78.6 &\textbf{ 86.8 }& 98.3 & 94.7 &\textbf{ 100.0} & 90.0 & 84.4 & \textbf{97.3 } \\ 
guard &81.9 & 80.2 &\textbf{ 88.5 }& 94.3 & 90.6 & \textbf{97.5 }& 89.4 & 87.2 & \textbf{95.0 } \\ 
guitarist &75.9 & 79.3 & \textbf{79.5} & 90.5 & 93.7 & \textbf{95.1} & 80.0 & 84.4 & \textbf{86.3 } \\ 
gymnast &\textbf{87.7 }& 85.5 & - &\textbf{ 96.2} & 95.6 & - &\textbf{ 92.4} & 89.9 & -  \\ 
hairdresser &76.8 & \textbf{79.4} & 79.0 & 94.1 & \textbf{96.9} & 92.9 & 82.4 & 79.9 &\textbf{ 83.3 } \\ 
horseman &\textbf{70.9 }& 62.1 & 64.5 & \textbf{85.4} & 75.7 & 80.0 & \textbf{77.2} & 67.9 & 70.0  \\ 
judge &25.7 &\textbf{ 31.3} & 28.3 & 28.6 & \textbf{35.3} & 33.3 & 28.6 &\textbf{ 33.8} & 33.3  \\ 
laborer &\textbf{75.3} & 73.1 & 74.4 & 88.4 & 85.8 & \textbf{86.1 }& \textbf{79.9 }& 78.9 & \textbf{79.9 } \\ 
lawman &\textbf{71.5} & 70.1 & 67.1 & \textbf{79.0} & 77.7 & 74.3 &\textbf{ 75.5 }& 74.6 & 70.6  \\ 
lifeguard &41.8 & 46.1 &\textbf{ 52.5} & 51.7 & 54.9 & \textbf{62.5 }& 47.5 & 49.8 & \textbf{62.5  }\\ 
machinist &\textbf{60.0} & 49.9 & 41.1 & \textbf{63.9} & 56.5 & 44.4 & \textbf{63.9} & 52.2 & 44.4  \\ 
motorcyclist &\textbf{57.9} & 52.7 & 51.9 & \textbf{81.6} & 78.2 & 69.2 & \textbf{60.9} & 54.2 & 57.7  \\ 
nurse &\textbf{83.4 }& 81.5 & 81.7 & \textbf{95.6} & 93.9 & 91.3 &\textbf{ 90.5 }& 86.1 & 82.6  \\ 
painter &54.0 & 58.9 & \textbf{68.6 }& 60.8 & 66.3 & \textbf{73.8 }& 58.2 & 62.3 & \textbf{73.8 } \\ 
patient &64.1 & 66.9 & \textbf{67.1} & \textbf{87.0 }& 85.6 & 86.5 & 65.6 & \textbf{69.2 }& 68.3  \\ 
prayer &82.8 & 83.0 & \textbf{85.2} & \textbf{96.0 }& 95.2 & 95.2 & 89.0 &\textbf{ 89.5 }&\textbf{ 89.5}  \\ 
referee &70.2 & 77.5 & \textbf{84.9} & 75.5 & 85.3 & \textbf{91.4 }& 73.6 & 80.9 &\textbf{ 88.6 } \\ 
repairman &\textbf{71.2 }& 61.7 & 65.2 & \textbf{77.6} & 69.7 & 71.0 & \textbf{75.0} & 65.5 & 69.6  \\ 
reporter &21.7 & 22.9 &\textbf{ 25.0} & 23.7 & 25.7 &\textbf{ 29.2 }& 22.4 & 23.7 &\textbf{ 25.0 } \\ 
retailer &33.3 & 35.0 &\textbf{ 52.2} & 40.9 & 43.2 & \textbf{59.5 }& 33.6 & 38.6 &\textbf{ 54.1 } \\ 
runner &90.9 & 85.9 & \textbf{91.1} & 99.2 & 95.2 & \textbf{100.0 }& 97.7 & 90.6 & \textbf{100.0 } \\ 
sculptor &74.5 & 73.0 &\textbf{ 85.0} & 81.8 & 82.4 & \textbf{95.8 }& 77.3 & 77.6 &\textbf{ 83.3 } \\ 
seller &73.0 & 73.2 &\textbf{ 74.5} & 87.4 & 87.0 &\textbf{ 88.5} & \textbf{82.0 }& 79.0 & 80.8  \\ 
singer &80.6 & 80.9 &\textbf{ 85.0} & 88.8 & 88.2 &\textbf{ 96.1 }& 85.0 & 85.5 & \textbf{88.2}  \\ 
skateboarder &40.7 & \textbf{43.1 }&- & 45.1 & \textbf{46.5} & - & 43.4 & \textbf{46.1} & -  \\ 
soccer player &81.8 & \textbf{82.9} & - & 90.6 & \textbf{91.7} & - & 86.2 &\textbf{ 87.1 }& -  \\ 
soldier &\textbf{65.1} & 63.4 & 51.4 &\textbf{ 72.8 }& 72.7 & 56.8 & \textbf{69.7 }& 68.3 & 54.1  \\ 
speaker &83.0 & 80.8 & \textbf{85.1 }& 89.3 & 88.5 & \textbf{93.1} & 87.7 & 85.6 &\textbf{ 89.1 } \\ 
student &60.6 & \textbf{71.1 }& -& 69.8 & \textbf{80.9} & - & 64.6 & \textbf{74.4} & - \\ 
teacher &\textbf{83.4 }& 81.0 & 80.0 &\textbf{ 96.6} & 90.3 & 87.5 & \textbf{93.1 }& 85.8 & 87.5  \\ 
tennis player &\textbf{94.2} & 93.8 & - & \textbf{98.9} &\textbf{ 98.9 }& - & 97.2 & \textbf{97.8} & - \\ 
trumpeter &22.8 & 29.5 & \textbf{38.4} & 26.7 & 34.8 & \textbf{45.5} & 25.6 & 31.4 & \textbf{38.2}  \\ 
waiter &76.2 & \textbf{77.6 }&- & 92.4 & \textbf{92.9} & - & \textbf{83.3} & 82.5 & - \\

\hline
avg & 64.6 & 64.0  & \textbf{ 68.2 } &   74.1 &  74.4  &  \textbf{76.4 }& 68.6  &  67.9 & \textbf{  72.4 }  \\       
    \end{tabular}
    \subcaption{Results for Detic}
\end{subtable}
\hspace*{3em}
\scriptsize
\begin{subtable}[t]{0.44\textwidth}
    \begin{tabular}{r|ccc | ccc | ccc}
        & \multicolumn{3}{c}{\small mAR}
        & \multicolumn{3}{c}{\small AR$_{0.5}$}
        & \multicolumn{3}{c}{\small AR$_{0.75}$}\\
        & \rotatebox{90}{\footnotesize young} & \rotatebox{90}{\footnotesize middle} & \rotatebox{90}{\footnotesize older}
        & \rotatebox{90}{\footnotesize young} & \rotatebox{90}{\footnotesize middle} & \rotatebox{90}{\footnotesize older}
        & \rotatebox{90}{\footnotesize young} & \rotatebox{90}{\footnotesize middle} & \rotatebox{90}{\footnotesize older}
        \\\hline

  & 0.0 & \textbf{30.3} & - & 0.0 & \textbf{51.4} & - & 0.0 & \textbf{32.9 }& - \\ 
& 7.7 & 5.9 & \textbf{11.5} & 15.5 & 11.2 & \textbf{22.0 }& 6.8 & 5.8 & \textbf{9.8} \\ 
& 43.9 & 38.0 &\textbf{ 58.0} & 72.7 & 67.2 & \textbf{100.0} & 50.6 & 41.5 & \textbf{80.0} \\ 
& 0.0 & \textbf{12.5} & - & 0.0 & \textbf{25.0} & - & 0.0 & \textbf{8.3} & - \\ 
& 26.2 & \textbf{24.2} & - & \textbf{46.4} & 41.3 & - & 25.0 & \textbf{26.0 }& - \\ 
& 1.4 & \textbf{1.9} & 1.5 & 4.5 & 6.2 & \textbf{8.1 }& 0.6 & \textbf{0.9 }& 0.0 \\ 
& 0.0 & \textbf{2.4} & 0.0 & 0.0 & \textbf{5.6} & 0.0 & 0.0 & \textbf{2.2 }& 0.0 \\ 
& \textbf{20.0} & 12.0 & - & \textbf{41.9 }& 20.0 & - & \textbf{16.1} & 15.0 & - \\ 
& 0.0 & \textbf{1.2} & 0.0 & 0.0 &\textbf{ 2.5} & 0.0 & 0.0 & \textbf{0.8 }& 0.0 \\ 
& 7.9 & 5.9 & \textbf{8.6 }& 12.9 & 10.3 &\textbf{ 14.3 }& \textbf{8.2} & 6.2 & 7.1 \\ 
& 33.8 & 39.1 & \textbf{40.7} & 55.2 & 62.6 & \textbf{66.9 }& 37.9 & 45.7 & \textbf{47.6} \\ 
& \textbf{37.6} & 32.0 & 24.3 & \textbf{68.8} & 55.7 & 57.1 & \textbf{39.0} & 37.1 & 28.6 \\ 
& 3.5 & \textbf{3.8} & - & \textbf{6.8} & 6.5 & - &\textbf{ 4.1} & 3.0 & - \\ 
& 33.6 & 30.9 & \textbf{38.0} & 55.2 & 52.1 &\textbf{ 60.8 }& 40.2 & 33.6 & \textbf{45.1} \\ 
& \textbf{5.0 }& 3.8 & 0.7 &\textbf{ 9.2} & 8.1 & 1.7 & \textbf{4.6 }& 3.3 & 0.0 \\ 
& 0.0 & \textbf{1.2} & 0.0 & 0.0 &\textbf{ 1.6 }& 0.0 & 0.0 & \textbf{1.6} & 0.0 \\ 
& 6.2 & 5.0 & \textbf{6.6} & 12.8 & 9.9 & \textbf{13.7 }& 5.1 & 4.4 & \textbf{5.5} \\ 
& 14.0 & 14.7 &\textbf{ 22.0 }& 26.7 & \textbf{32.9} & 60.0 & 13.3 & 12.5 & \textbf{20.0} \\ 
& \textbf{15.0} & 10.5 & 11.7 & \textbf{31.8 }& 19.9 & 20.8 & 9.1 & 9.9 & \textbf{12.5} \\ 
& 11.9 & 18.3 & \textbf{27.9} & 32.6 & 40.1 & \textbf{58.1} & 7.0 & 14.6 & \textbf{24.2 }\\ 
& 14.1 & 15.2 & \textbf{19.2} & 34.0 & 31.9 & \textbf{38.5} & 9.6 & \textbf{12.5} & 11.5 \\ 
& 19.8 & 19.6 & \textbf{32.0} & 38.7 & 35.5 & \textbf{56.7 }& 18.5 & 20.6 & \textbf{36.7 }\\ 
& \textbf{10.0} & 8.5 & -  & \textbf{19.1} & 17.1 & -  & \textbf{9.8} & 9.1 & -  \\ 
& \textbf{15.2} & 13.3 & 12.1 & \textbf{28.0} & 24.3 & 23.5 & 12.0 & 12.8 & \textbf{14.7 }\\ 
& 13.4 & \textbf{14.5} & 11.0 & 36.5 & \textbf{38.8} & 30.0 & 3.8 &\textbf{ 5.4} & 0.0 \\ 
& -- & \textbf{10.4} & 0.0 & -- & \textbf{25.0} & 0.0 & -- & \textbf{3.6} & 0.0 \\ 
& 23.1 & 21.9 & \textbf{28.9} & 44.0 & 46.2 & \textbf{58.6} & 22.0 & 17.4 & \textbf{24.3} \\ 
& 20.2 & 21.1 & \textbf{22.6 }& 42.0 & 43.1 & \textbf{46.2 }& 18.3 & 18.8 & \textbf{21.5} \\ 
& \textbf{7.5 }& 7.0 & 0.0 & \textbf{19.7} & 17.9 & 0.0 & 2.8 & \textbf{5.2 }& 0.0 \\ 
& 21.7 & 21.3 & \textbf{23.3} & 34.8 & 35.5 & \textbf{41.7 }& \textbf{26.1} & 25.0 & 25.0 \\ 
&\textbf{ 21.9 }& 15.5 & 19.2 & \textbf{50.0} & 37.0 & 37.5 & 12.3 & 9.6 & \textbf{20.8} \\ 
& 31.8 & 24.8 & \textbf{34.5 }& \textbf{52.2} & 43.6 & 50.0 & 37.2 & 26.0 &\textbf{ 40.0} \\ 
& \textbf{18.0 }& 15.6 & 17.7 & \textbf{30.1} & 29.3 & 27.6 & \textbf{23.3} & 16.1 & 20.4 \\ 
& \textbf{28.5} & 26.5 & 26.6 &\textbf{ 50.3} & 47.6 & 45.2 & \textbf{29.7} & 27.6 & 28.0 \\ 
& 0.0 & 2.7 & \textbf{2.8} & 0.0 & \textbf{5.5} & 4.3 & 0.0 & 1.8 & \textbf{2.9 }\\ 
& 19.6 & 20.4 & \textbf{21.4} & 40.8 & 40.1 & \textbf{45.7 }& 16.3 & 19.8 &\textbf{ 22.9} \\ 
& \textbf{20.1} & 17.9 & 17.0 & \textbf{39.5 }& 32.8 & 30.4 & 18.4 & \textbf{19.1} & 17.4 \\ 
&\textbf{ 9.2} & 5.2 & 4.5 & \textbf{19.7} & 13.0 & 6.9 & \textbf{7.0} & 3.9 & 3.4 \\ 
& 1.0 & 2.5 &\textbf{ 3.1} & 2.8 & 6.5 & \textbf{6.9 }& 0.0 & 1.5 & \textbf{3.4 }\\ 
& 7.5 & \textbf{8.3 }& 0.0 &\textbf{ 21.6} & 21.1 & 0.0 & 3.9 & \textbf{4.7} & 0.0 \\ 
& \textbf{2.4} & 2.3 & 0.0 & \textbf{5.9 }& 5.1 & 0.0 & 0.0 & \textbf{3.1} & 0.0 \\ 
& 7.8 & 8.2 & \textbf{9.8} & 16.8 & 16.6 & \textbf{21.1} & 6.2 & 6.9 & \textbf{8.8} \\ 
& \textbf{5.1} & 3.9 & 1.7 & \textbf{10.3} & 7.3 & 5.8 & \textbf{4.5} & 3.6 & 0.0 \\ 
& 21.5 & \textbf{23.9} & -- & 46.9 & \textbf{49.7} & -- & 18.8 & \textbf{22.1} & -- \\ 
& \textbf{26.7 }& 22.6 & -- & \textbf{49.1 }& 42.8 & -- & \textbf{24.6} & 21.4 & -- \\ 
& \textbf{16.3} & \textbf{16.3} & 1.2 & \textbf{40.0} & 33.8 & 6.2 & 9.2 & \textbf{14.5} & 0.0 \\ 
& 2.0 & 1.7 & \textbf{2.1 }&\textbf{ 4.6 }& 3.2 & 3.7 & \textbf{1.9 }& 1.7 & 1.6 \\ 
& \textbf{29.0 }& 25.3 & 0.0 &\textbf{ 51.9} & 44.7 & 0.0 & \textbf{33.8} & 25.0 & 0.0 \\ 
& \textbf{28.1} & 22.2 & 15.0 & \textbf{51.6 }& 39.8 & 50.0 & \textbf{29.0 }& 24.8 & 0.0 \\ 
& 32.5 & \textbf{33.8} & - & 60.0 & \textbf{62.2} &- & 32.7 & \textbf{34.4} & - \\ 
& \textbf{5.3} & 5.1 & 3.6 &\textbf{ 11.6 }& 10.3 & 5.1 & 2.3 &\textbf{ 5.7 }& 5.1 \\ 
&\textbf{ 5.2 }& 4.2 & - & \textbf{10.4} & 8.6 & - & \textbf{4.2 }& 4.0 & - \\ \hline
  & \textbf{14.7} & 14.5 & 14.0 & \textbf{28.5} & 27.8 & 26.2 & 13.8 & 14.1 & \textbf{15.0}
        
    \end{tabular}
\subcaption{Results for OFA}
\end{subtable}
\vspace{2em}
\caption{The average recall (AR) results for Detic (detection) and OFA (visual grounding) across the 52 person-related classes for each \age label. The  highest recall numbers are bolded.}
\label{tab:age_both}
\end{table*}

\subsubsection{Visual Grounding}\label{sec:exp_setup_ofa}
We evaluated OFA \cite{wang2022unifying}.
For OFA, we used the pretrained version OFA$_{large}$ in the HuggingFace Transformers library \cite{wolf2020transformers}; we did not perform any additional finetuning.
We used beam-search with 5 beams, top-p=0.6 and limited the generation to a maximum of 100 new tokens.
We prompted OFA with the input (\eg ``Which region does the text \texttt{\{person class\}} describe?").
Because OFA produces a single bounding box per prompt, we only evaluated images that contained no more than one person instance per person class.
7858 images were excluded because they contained multiple instances per class.
We show the average recall across different IoUs and for different \age labels in Table \ref{tab:age_both}.

\clearpage
\onecolumn
\section{Data Card}
We provide a data card for FACET, following the guidance of \cite{holland2020dataset}.
\begin{longtable}{|p{5cm}|p{11cm}|}

\endfirsthead
\hline
\multicolumn{2}{|c|}{\textbf{FACET Data Card}}\\
\hline
\endhead
\hline
\endfoot
\hline
\endlastfoot
\hline
\multicolumn{2}{|c|}{}\\
\multicolumn{2}{|c|}{\textbf{\large FACET}}\\*
\multicolumn{2}{|c|}{\facetlink}\\
\multicolumn{2}{|c|}{}\\
\multicolumn{2}{|c|}{\parbox{17cm}{\small{
FACET is a large, publicly available evaluation set of \datasetsize images for the most common vision problems - \textbf{image classification}, \textbf{object detection}, \textbf{segmentation} and \textbf{visual grounding}. People in FACET are annotated with person-related attributes such as \textbf{\skintone} and \textbf{hair type}, \textbf{bounding boxes} and labeled with fine-grained \textbf{person-related classes} such as \textit{disk jockey} or \textit{guitarist}.
}}}\\
\multicolumn{2}{|c|}{}\\*

\hline
\multicolumn{2}{|c|}{\textbf{Overview}} \\*
\hline
\vspace{0.1em} 
Publisher & \vspace{0.1em}  Meta AI Research, FAIR \vspace{0.8em}  \\*
Authors & Laura Gustafson, Chloe Rolland,  Nikhila Ravi, Quentin Duval, Aaron Adcock, Cheng-Yang Fu, Melissa Hall, Candace Ross \vspace{0.8em}   \\ 
Contact & \href{mailto:facet@meta.com}{facet@meta.com} \vspace{0.8em}  \\ 
Funding \& Funding Type & Industry\vspace{0.8em}  \\ 
License & Custom license, see dataset download agreement \vspace{0.1em} \\*
\hline

\multicolumn{2}{|c|}{\textbf{Applications}} \\*
\hline
\vspace{0.2em} 
Dataset Purpose & 
\vspace{0.2em} 
Evaluate computer vision models to detect potential fairness concerns \vspace{0.8em}  \\*
Key Application &
\textit{Computer Vision, Fairness and Robustness}\vspace{0.8em}   \\*
Primary Motivations & 
Give researchers a tool to help understand model fairness.
Allow researchers to investigate how the demographic attributes of a person in the image correlates with model performance. FACET supports common vision tasks, with annotations for classification detection, and segmentation. \vspace{0.8em}  \\
Intended Audience &  
Researchers aiming to detect potential fairness concerns and biases in their trained vision models.\vspace{0.8em}   \\*
Suitable Use Case & 
\textbf{FACET is for evaluation only.} \vspace{0.1em} \\*
\hline
\multicolumn{2}{|c|}{\textbf{Data Type}}\\*
\hline
\vspace{0.5em} 
Primary Data Type & 
\vspace{0.5em} 
Images\vspace{0.8em}  \\*
Primary Annotation Types & Manually gathered annotations for:
\begin{itemize}
    \item Bounding boxes
    \item Category labels for the bounding boxes
    \item A series of demographic, robustness, and additional attributes for the person in the bounding box
    \item Manually annotated labels for masks from Segment Anything 1 Billion (SA-1B\cite{segment_anything}). These masks were automatically generated by the Segment Anything Model (SAM).
\end{itemize} \\  \noalign{\penalty-5000}
\vspace{0.5em} 
Data SnapShot  & \vspace{0.5em} 
\begin{itemize}
    \item \textbf{\datasetsize} images
    \item exhaustive annotations for \textbf{\peoplesize} people
    \item \textbf{\numpersonclasses} categories for people that include occupations, athletes, artists, etc
    \item \textbf{\attributenum} attributes annotated for person including demographic attributes such as \textit{\gender} robustness annotations such as \textit{lighting condition} and additional attributes such as \textit{hair color}
    \item \textbf{3} mask labels \textit{person, clothing, hair} for 69k masks. Masks and mask labels are not exhaustive. 17k people in 14k images have labelled masks. Additional unlabeled masks from \datasourcenocite\ are compatible with \datasetabbr.
\end{itemize}
\\ 
Data Sources & Images come from SA-1B. \newline \\

Annotation format & JSON files of COCO formatted\footnote{
\url{https://cocodataset.org/\#format-data}
} annotations for the bounding boxes and masks are provided, and person attribute annotations are given in a CSV file.  \vspace{0.1em} \newline

\texttt{coco\_boxes.json:} COCO-style JSON file containing bounding boxes for people in FACET. The id field for each annotation is the same as \texttt{person\_id}.
\texttt{Category\_id} corresponds to the primary class for the person. \vspace{0.1em} \newline

\texttt{coco\_masks.json:}  COCO-style JSON file containing Segment Anything Model (SAM) generated masks for people in FACET. Each annotation has an additional \texttt{person\_id} field, corresponding to the \texttt{person\_id} in \texttt{annotations.csv} and \texttt{coco\_boxes.json}. Categories for masks are one of [person, clothing, hair]. Masks are non-exhaustive. \vspace{0.1em} \newline

\texttt{annotations.csv:} A CSV containing the annotations per person is be provided. Each item in the annotation file contains:
\begin{enumerate}
    \item Reference information:
        \begin{itemize}
            \item \texttt{filename}
            \item \texttt{person\textunderscore id}: unique integer representing the annotation
        \end{itemize}
    \item Task information:
            \begin{itemize}
            \item \texttt{class1}: This is the primary category the person matches. \textit{Cannot be None.}
            \item \texttt{class2}: This is the secondary category the person matches. \textit{Can be None.}
            \item \texttt{bounding\textunderscore box}: Person bounding box.
        \end{itemize}
    \end{enumerate} \\
    &    \begin{enumerate}\addtocounter{enumi}{2}
        \item Demographic Attribute annotations.
        \begin{itemize}
            \item \textit{\gender}: All of the following annotations will given in a binary fashion: \texttt{[gender\textunderscore presentation\textunderscore masc,  \newline gender\textunderscore presentation\textunderscore non\textunderscore binary,
            \newline gender\textunderscore presentation\textunderscore fem, gender\textunderscore presentation\textunderscore na]}
            \item \textit{\skintone}: Each annotators annotations are considered per MST in a binary fashion. Annotations from all annotators are summed into a single value per MST, so the value at $MST_{i}$ may be greater than 1. Values will be given for all of the following: \texttt{[skin\textunderscore tone\textunderscore 1, \dots, skin\textunderscore tone\textunderscore 10,
            skin\textunderscore tone\textunderscore na]}
            \item \textit{\age}: all of the following annotations are included in a binary fashion: \texttt{[age\textunderscore presentation\textunderscore young, age\textunderscore presentation\textunderscore middle, age\textunderscore presentation\textunderscore older, age\textunderscore presentation\textunderscore na]}     
        \end{itemize}

    \item Additional Attribute information: All binary values.
        \begin{itemize}
        \item \textit{hair color:} \texttt{[hair\_color\_black, hair\_color\_red, \newline hair\_color\_blonde, hair\_color\_brown, hair\_color\_colored, hair\_color\_grey, hair\_color\_na]}
        \item \textit{hair type:} \texttt{[hair\_type\_wavy, hair\_type\_curly, \newline hair\_type\_coily, hair\_type\_straight, hair\_type\_bald, hair\_type\_dreadlocks, hair\_type\_na]}
        \item \textit{other items:} \texttt{[has\_eyewear, has\_headscarf, \newline has\_tattoo, has\_cap, has\_facial\_hair, has\_mask]}
    \end{itemize}
        \item Robustness Annotations: All binary values.
        \begin{itemize}
        \item \textit{lighting condition:} \texttt{[lighting\textunderscore overexposed, \newline
        lighting\textunderscore underexposed, lighting\textunderscore well\textunderscore lit, lighting\textunderscore dimly\textunderscore lit]}
        \item \textit{visibility:} \texttt{[visible\textunderscore minimal, visible\textunderscore torso, \newline visible\textunderscore face]}
        \end{itemize}

\end{enumerate}

\end{longtable}
\clearpage
\twocolumn

\section{FACET CrowdWorkSheets}
\label{app:crowdworksheets}
To further describe our annotation process, we answer the questions posed in CrowdWorkSheets\cite{D_az_2022}.
\subsection{Task Formulation}
\paragraph{At a high level, what are the subjective aspects of your task?}
Annotating the \textit{perceived} attributes of a person is by nature subjective. For perceived skin tone we expected the annotations would be subjective and have high variance. To account for this, we gather annotations from three annotators and release the cumulative results of all three. For subjectivity across the other attributes and labeling classes, we provided annotators with diverse representations of each attribute or class in the guidelines to try to minimize annotator bias.

\paragraph{What assumptions do you make about annotators? How did you choose the specific wording of your task instructions? What steps, if any, were taken to verify the clarity of task instructions and wording for annotators?}

To qualify for the annotation task, annotators had to pass a strong English requirement. For the annotation of perceived skin tone only, we had a more lenient English requirement to increase the diversity of the annotators, and additionally translated the annotation instructions into Spanish. 

As we were annotating images, we provided visual examples for all of the annotations and classes. We sourced multiple examples per attribute (\eg brown hair) and class (\eg doctor), with at least one example for someone with more stereotypical maleness with the attribute and someone with more stereotypical femaleness with the attribute. For classes, we sourced multiple examples of someone who would qualify for a given class (\textit{e.g for dancer we sourced images of both a ballerina and a break-dancer}). For given examples for the Monk Skin Tone (MST) scale, we sourced four examples per MST value, and attempted to capture some of the diversity within a specified MST value.

\paragraph{What are the precise instructions that were provided to annotators?}
The goal of the project is to build a dataset that helps determine if Computer Vision models have biases based on the apparent attributes of the person in the photo. We are creating an image classification dataset that also contains labels of the apparent protected attributes of the people in the image. The dataset is for evaluation only, and is to help better analyze and detect potential biases. The protected attributes will not in any way be used for training a model. We are not collecting any biometric information about the people in the photos. 

\begin{enumerate}
    \item  \textbf{Target category classification:} Given an image, and a target category, we aim to determine if the image is a good representation for the category. The annotators will mark whether or not there is a person in the photo matching the category, and if so if there are $\leq 5$ people who match this category. The categories will be all people related - such as doctor, soccer player, etc. Multiple images will be shown per task to annotate. The default response will be ‘No person matches this category’.
    \item \textbf{Bounding boxes and classification labels for people:}
Given an image, draw bounding boxes around all people who match any of the list of categories. For each bounding box around a person, mark which category they belong to. If they belong to multiple categories, you should mark the second category under ‘secondary category’.
\item \textbf{3. Apparent skin tone annotations}
Given an image, with a bounding box around a person, annotate the person’s apparent skin tone. You may select as many skin tones from the list as you feel appropriate. If it is not possible to tell the skin tone from the photo, please mark cannot be determined. Please select at least two values for the skin tone, and make sure that the values that you select are consecutive. If it is too hard to determine the annotation, mark the values it appears and cannot be determined.
Zoom in (option + mouse scroll) as necessary in order to determine the skin tone.
\item \textbf{4. Apparent attribute annotations}
Given an image, with a bounding box around a person, annotate the given apparent attributes of the person. For each category, see the examples given. If it is not possible to determine the attribute from the photo, please mark cannot be determined. 
Apparent lighting condition is on the person: Please indicate how the lighting is with respect to the person in the bounding box. If the lighting is between two categories, mark both. 
\end{enumerate}

\subsection{Selecting Annotations}
\paragraph{Are there certain perspectives that should be privileged? If so, how did you seek these perspectives out?}
No. N/A

\paragraph{Are there certain perspectives that would be harmful to include? If so, how did you screen these perspectives out?}
Harmful perspectives would include annotators who had a clear bias in their annotations. We screened these perspectives out by using training, and only including production raters who had high accuracy on the training set. Annotators with consistent bias would likely not have been able to get a high enough accuracy on the training to graduate.

\paragraph{Were sociodemographic characteristics used to select annotators for your task? If so, please detail the process. If you have any aggregated sociodemographic statistics about your annotator pool, please describe. Do you have reason to believe that sociodemographic characteristics of annotators may have impacted how they annotated the data? Why or why not? }
We sourced geographically diverse annotators from the following 7 countries during our annotation process: United States, Philippines, Egypt, Colombia, Taiwan, Spain and Kenya. The breakdown of annotators per region is shown in Figure \ref{fig:annotator_pie1} in the main text. 

\paragraph{If you have any aggregated sociodemographic statistics about your annotator pool, please describe. Do you have reason to believe that sociodemographic characteristics of annotators may have impacted how they annotated the data? Why or why not?
}
Other sociodemographic statistics about our annotator pool were not available.

\paragraph{Consider the intended context of use of the dataset and the individuals and communities that may be impacted by a model trained on this dataset. Are these communities represented in your annotator pool?}
The FACET benchmark is to be used for evaluation purposes only. The underlying images in FACET are geographically diverse. To incorporate geographic diversity into our annotation process, we sourced annotators from 7 countries across regions.

\subsection{Platform and Infrastructure Choices}

\paragraph{What annotation platform did you utilize? At a high level, what considerations informed your decision to choose this platform? Did the chosen platform sufficiently meet the requirements you outlined for annotator pools? Are any aspects not covered?}
We used a proprietary annotation platform.

\paragraph{What, if any, communication channels did your chosen platform offer to facilitate communication with annotators? How did this channel of communication influence the annotation process and/or resulting annotations?}
For Stage 2 (drawing and labeling bounding boxes for person classes), annotators' annotations were compared to a golden set and were required to achieve IoU above 85\% to pass. After these training stages, annotations were manually reviewed and the annotators were given feedback for improvement. Following this, if annotators had high quality labels when spot-checked, they graduated to annotating images for the final benchmark.

We provided annotators individualized feedback during their training for drawing bounding boxes on a daily basis. Our vendor communicated to annotators common types of mistakes that we witnessed during training, and the corresponding corrections.

\paragraph{How much were annotators compensated? Did you consider any particular pay standards, when determining their compensation? If so, please describe.}
Annotators were compensated with an hour wage set per country.

\subsection{Dataset Analysis and Evaluation}
\paragraph{How do you define the quality of annotations in your context, and how did you assess the quality in the dataset you constructed?}
For each task, annotators were first placed into training for the task.
They were asked to annotate a large number of examples per task. We hand annotated the same examples and used our annotations as the ground truth measured the accuracy per annotator. Annotators were graduated from training when their accuracy reached above a given threshold. For the task requiring annotators to draw bounding boxes around people, annotators were only graduated after we manually spot checked the their bounding boxes to ensure quality. During the perceived skin tone annotation task, we asked annotators if they agreed with the class label, and grade the quality of the given bounding box. If one of the three annotators disagreed with the class label or bounding box, the annotation was removed, and the image added to the queue of images for task 2 (drawing bounding boxes).

\paragraph{Have you conducted any analysis on disagreement patterns? If so, what analyses did you use and what were the major findings?}
We pointed out common mistakes during weekly meetings with the vendor. While in training, we noticed consistent mistakes among annotators that we corrected before graduation. The most common mistake was around drawing the bounding boxes: many annotators during training would draw bounding boxes that included objects the person was holding \textit{e.g guitar}. With the weekly meetings and individualized feedback, we were able to address this.

\paragraph{How do the individual annotator responses relate to the final labels released in the dataset?}
For perceived skin tone only, we sourced three annotations per person in the dataset. We release the annotations from all three annotators, giving a distribution over perceived skin tone per person in the dataset. We believe that a distribution more accurately describes a person’s perceived skin tone than a single value.

\subsection{Dataset Release and Maintenance}
\paragraph{Do you have reason to believe the annotations in this dataset may change over time? Do you plan to update your dataset?
}
At this time we do not plan to have updates for this dataset. We will allow users to flag any images that may be objectionable content, and remove objectionable content if found.

\paragraph{Are there any conditions or definitions that, if changed, could impact the utility of your dataset?}
The FACET benchmark contains examples for many different types of professions, athletes, artists, etc. If over time the way these occupations look shifts, this could impact the dataset. As a concrete example, there are a number of images in the dataset that were taken since the beginning of the COVID-19 pandemic. Many doctors and nurses in the dataset are wearing much more PPE than in images of doctors and nurses from before the COVID-19 pandemic.

\paragraph{Will you attempt to track, impose limitations on, or otherwise influence how your dataset is used? If so, how?}

The FACET benchmark is for evaluation purposes ONLY. Using FACET annotations for training is strictly prohibited. Users must agree to the terms of use before downloading the dataset.

\paragraph{Were annotators informed about how the data is externalized? If changes to the dataset are made, will they be informed?} No. No.

\paragraph{Is there a process by which annotators can later choose to withdraw their data from the dataset? If so, please detail.}
No.

\onecolumn
\section{Fine-grained dataset statistics}

\begin{table*}[h!]
\scriptsize
\begin{center}
\begin{tabular}{|l|r|p{2.5em}p{2em}p{1.5em}|rrrrrrrrrr|rrr|}
\hline

 & \multicolumn{17}{|c|}{\makecell{\vspace{0.1em}\\ \large FACET Class Statistics \\ \vspace{0.1em}}} \\
\hline
 & & \multicolumn{3}{c|}  {\makecell[c]{    Perceived Gender\\
    Presentation}}
 &\multicolumn{10}{c|}{\skintoneallcap }
  &\multicolumn{3}{c|}{\makecell[c]{
    Perceived Age\\
    Group}} \\
Person Class&Total& \rotatebox{90}{\scalebox{1.1}{\parbox{1.25cm}{\tiny{stereotypical\\maleness}}}} &\rotatebox{90}{\scalebox{1.1}{\parbox{1.25cm}{\tiny{stereotypical\\femaleness}}}}

&\rotatebox{90}{\scalebox{1.1}{\parbox{1.25cm}{\tiny{non-binary\\presentation}}}}

&1 \mstone & 2 \msttwo & 3 \mstthree
 & 4 \mstfour & 5 \mstfive & 6 \mstsix 
 & 7 \mstseven & 8 \msteight & 9 \mstnine & 10 \mstten &\rotatebox{90}{younger}&\rotatebox{90}{middle}&\rotatebox{90}{older}\\
\hline
\hline
lawman&4609&3768&403&3&560&2363&2881&2642&1825&1215&615&322&166&74&387&3151&144 \\
laborer&3030&2208&378&1&112&577&887&1171&1171&1269&844&508&291&136&297&1643&193 \\
boatman&2147&1074&742&5&137&741&991&995&794&573&296&134&86&31&482&906&147 \\
guard&1851&1597&121&4&306&1045&1208&1047&714&470&246&119&62&27&317&1181&48 \\
backpacker&1738&1006&458&4&167&771&1010&951&761&475&193&93&50&28&361&842&53 \\
basketball player&1680&1479&134&2&307&869&991&845&632&461&413&359&332&165&492&1056&3 \\
tennis player&1663&1058&488&0&147&805&1152&1262&1002&617&234&126&90&57&360&1081&6 \\
farmer&1632&823&539&1&50&208&335&466&635&816&681&450&216&87&129&844&227 \\
soldier&1561&1336&75&0&204&766&892&802&578&463&281&130&66&22&237&972&39 \\
singer&1518&1013&428&14&240&824&1013&931&677&399&184&140&93&46&357&984&89 \\
dancer&1475&510&812&10&207&644&863&798&716&431&214&113&67&26&567&644&32 \\
speaker&1470&1119&282&1&152&789&1093&1050&755&392&182&92&49&25&134&987&207 \\
motorcyclist&1468&822&302&4&55&384&518&583&539&478&249&104&59&22&213&577&56 \\
repairman&1430&1187&54&1&65&480&681&815&753&676&373&150&60&18&126&836&78 \\
seller&1342&699&533&6&74&359&555&705&757&642&379&192&74&28&205&758&184 \\
ballplayer&1316&1145&62&1&104&584&743&781&635&473&253&168&112&51&214&834&12 \\
guitarist&1279&1115&87&3&138&678&843&816&596&330&139&73&50&26&233&802&116 \\
computer user&1267&597&322&2&176&641&818&785&608&358&152&71&34&12&258&449&24 \\
soccer player&1233&1102&34&1&113&521&692&711&559&364&200&126&128&76&322&732&5 \\
craftsman&1127&785&220&4&75&321&467&598&631&627&389&210&92&36&117&599&188 \\
nurse&1124&322&535&3&115&368&505&536&529&399&188&86&27&6&169&547&24 \\
drummer&1006&744&162&3&114&428&534&483&388&331&222&150&99&42&256&530&68 \\
skateboarder&1000&818&88&1&82&468&635&650&463&281&136&62&37&14&360&465&1 \\
painter&983&590&251&0&77&318&460&530&506&420&246&123&56&22&168&435&129 \\
fireman&933&674&34&0&68&270&358&391&237&192&77&22&10&7&55&512&14 \\
patient&896&408&275&0&75&280&389&472&486&444&242&102&41&16&131&368&127 \\
horseman&884&491&290&1&152&538&592&484&287&127&54&26&12&5&181&512&22 \\
doctor&861&361&313&1&86&343&450&462&410&284&145&69&21&6&107&441&43 \\
prayer&810&444&265&3&58&223&307&355&394&357&195&99&51&18&104&358&124 \\
referee&776&694&38&1&88&417&539&547&374&186&80&38&23&10&54&584&35 \\
student&747&379&247&1&92&241&322&365&367&316&163&93&62&26&319&264&5 \\
runner&654&469&117&3&88&320&415&379&291&126&45&23&27&16&134&403&19 \\
gymnast&635&252&316&1&116&348&424&366&322&145&43&28&17&6&300&233&2 \\
retailer&561&296&234&0&53&198&301&298&298&196&90&41&14&5&114&332&39 \\
climber&551&355&92&2&59&231&306&301&251&155&73&26&13&8&107&261&4 \\
trumpeter&530&451&36&3&63&308&336&304&212&145&74&41&37&18&89&316&56 \\
lifeguard&529&398&62&0&20&160&232&286&229&186&103&52&28&8&118&273&8 \\
electrician&505&415&7&0&9&100&140&188&182&175&101&52&35&9&47&270&9 \\
gardener&499&266&173&1&45&187&257&265&235&197&108&56&33&16&66&245&79 \\
reporter&473&302&145&1&75&281&324&269&204&116&50&18&13&4&77&302&24 \\
hairdresser&461&342&85&3&32&143&209&257&242&237&145&75&35&17&69&294&43 \\
machinist&413&329&30&0&33&173&223&252&191&168&89&34&20&7&42&241&20 \\
cheerleader&410&78&314&0&77&191&292&268&205&88&38&18&12&3&246&117&5 \\
waiter&350&204&109&1&34&184&245&220&177&120&51&24&18&7&68&224&7 \\
disk jockey&318&228&27&1&43&162&200&194&127&77&37&27&20&10&67&167&2 \\
flutist&312&247&41&0&38&152&192&184&154&118&77&43&16&4&50&189&32 \\
astronaut&289&165&14&0&15&72&89&78&58&18&2&0&0&2&5&158&2 \\ 
carpenter&268&230&7&0&11&82&124&147&129&131&87&52&25&9&20&160&27 \\
sculptor&240&187&21&0&10&76&104&120&107&107&78&50&24&5&24&144&27 \\
teacher&216&116&76&1&28&104&141&142&108&76&36&16&10&4&31&150&9 \\
judge&101&67&28&0&11&50&76&71&44&21&6&3&1&0&8&71&12 \\
bartender&57&37&14&0&5&27&42&36&29&19&7&3&1&1&7&41&1 \\

\hline

\end{tabular}
\caption{\label{table:class_all}Number of people for each person class and demographic group in FACET.}
\end{center}
\end{table*}

\begin{table}[h!tb]
\small
\begin{centering}
\begin{tabular}{|l|rrr|}
\hline

 \multicolumn{4}{|c|}{\makecell{ \vspace{-0.6em} \\ \large FACET Mask Statistics\\ \vspace{-1em} }} \\
\hline
{} &  person &  clothing &  hair \\
\hline
\hline
\gender & & & \\ 
\hline
with stereotypical maleness      &      6608 &     32103 &  3788 \\
with stereotypical femaleness  &     4127 &     18136 &  3346 \\
non-binary presentation &      50 &       223 &    36 \\
cannot be determined         &     72 &       193 &    13 \\
\hline
\hline
\skintone & & & \\ 
\hline
MST 1 \mstone &   2198 &     10687 &  1389 \\
MST 2 \msttwo  &    5154 &     24328 &  3496 \\
MST 3 \mstthree  &  6121 &     28825 &  4263 \\
MST 4 \mstfour  &   5651 &     26583 &  3889 \\
MST 5 \mstfive  & 4849 &     22738 &  3349 \\
MST 6 \mstsix  &   3816 &     17931 &  2452 \\
MST 7 \mstseven & 2542 &     11845 &  1544 \\
MST 8 \msteight  &   1619 &      7564 &   922 \\
MST 9 \mstnine &  1216 &      5727 &   666 \\
MST 10 \mstten &     521 &      2481 &   293 \\
cannot be determined &     2839 &     11844 &  1611 \\
\hline
\hline
\age & & & \\
\hline
younger  &  4145 &     19440 &  3107 \\
middle &    5443 &     25458 &  3319 \\
older  &     1134 &      5352 &   733 \\
cannot be determined     &     135 &       405 &    24 \\
\hline
\hline
Hair color & & & \\
\hline
black   & 4053 &     18137 &  3323 \\
brown   &   2726 &     12205 &  2267 \\
blonde  &    1024 &      4633 &   952 \\
red/orange     &     148 &       674 &   136 \\
colored &   84 &       340 &    96 \\
grey    &   747 &      3519 &   559 \\
cannot be determined     &     2885 &     14863 &   485 \\
\hline
\hline
Hair type & & & \\
\hline

wavy       &   2090 &      9526 &  1897 \\
curly      &   241 &      1141 &   253 \\
straight   &   5141 &     22109 &  4395 \\
coily      &     178 &       750 &   158 \\
dreadlocks &  113 &       522 &   109 \\
bald       & 265 &      1167 &    81 \\
Unknown         &  3626 &     19129 &   905 \\
\hline
\hline
Additional attribute & & & \\
\hline
eyeware     &    1509 &      6993 &   957 \\
headscarf   &   665 &      3634 &   256 \\
tattoo      &  184 &       926 &   143 \\
cap         &   3305 &     18209 &   797 \\
facial\ hair &    1511 &      7382 &   963 \\
mask        &    591 &      3271 &   377 \\

\hline

\end{tabular}
\caption{\label{app:mask_stats}Number of masks per type for the attributes in FACET. For \skintonenospace, hair color, hair type, and additional attributes, a person in FACET can be marked with multiple values; therefore, the sum of the masks over the group of attributes may be greater than the total number of masks.}
\end{centering}
\end{table}

\end{appendix}

\end{document}